%% file: AverageMDP.tex
\documentclass[english]{article}

\usepackage{geometry}
\usepackage[T1]{fontenc}
\usepackage[latin9]{inputenc}
\usepackage{bm}
\usepackage{amsmath,mathtools}
\usepackage{amssymb}
\usepackage[unicode=true,
 bookmarks=false,
 breaklinks=false,pdfborder={0 0 1},colorlinks=false]
 {hyperref}
\hypersetup{
colorlinks,citecolor=blue,filecolor=blue,linkcolor=blue,urlcolor=blue}
\usepackage{xcolor,colortbl}
\definecolor{Gray}{gray}{0.85}

\makeatletter

\usepackage{amsthm}
\usepackage{cite}  
\usepackage{comment}
\usepackage{natbib}
\usepackage{booktabs}
\usepackage{graphicx}
\usepackage{enumitem}
\usepackage[linesnumbered,ruled,vlined]{algorithm2e}
\usepackage[noend]{algorithmic}

\usepackage{float}
\usepackage{multirow}
\usepackage{dsfont}
\usepackage{tcolorbox}
\usepackage{changepage}
\usepackage{color}
\definecolor{jiw}{RGB}{242,81,0} 
\definecolor{yjc}{RGB}{125,0,0} 
\definecolor{gen}{RGB}{200,0,0}

\newcommand{\note}[1]{\textcolor{black}{#1}}

\allowdisplaybreaks

\DeclareFontFamily{U}{matha}{\hyphenchar\font45}
\DeclareFontShape{U}{matha}{m}{n}{
<5>matha5<6>matha6<7>matha7<8>matha8<9>matha9
<10><10.95>matha10
<12><14.4><17.28><20.74><24.88>matha12
}{}
\DeclareSymbolFont{matha}{U}{matha}{m}{n}
\DeclareFontSubstitution{U}{matha}{m}{n}
\DeclareMathSymbol{\ovoid}{\mathbin}{matha}{"6C}
\renewcommand{\sp}[1]{\|{#1}\|_\mathsf{sp}}

\newcommand{\defn}{\coloneqq}

\newcommand{\cS}{\mathcal{S}}
\newcommand{\cA}{\mathcal{A}}

\newcommand{\mymid}{\,|\,}

\newcommand{\synset}{\mathcal{C}}
\newcommand{\syn}{\iota}

\definecolor{yuchen}{RGB}{0,200,100}

\date{\today}

\makeatother

\begin{document}

\theoremstyle{plain} \newtheorem{lemma}{\textbf{Lemma}}\newtheorem{proposition}{\textbf{Proposition}}\newtheorem{theorem}{\textbf{Theorem}}\newtheorem{assumption}{\textbf{Assumption}}\newtheorem{definition}{\textbf{Definition}}

\theoremstyle{remark}\newtheorem{remark}{\textbf{Remark}}

\title{Sample Complexity of Average-Reward Q-Learning: \\ From Single-agent to Federated Reinforcement Learning}

 \author{
    Yuchen Jiao\footnote{The first two authors contributed equally, and are listed in alphabetical order.} \thanks{Department of Statistics and Data Science, Chinese University of Hong Kong.} \\
    CUHK 
    \and
 	Jiin Woo\footnotemark[1] \thanks{Department of Electrical and Computer Engineering, Carnegie Mellon University.} \\
	CMU 
 	\and
    Gen Li\footnotemark[2] \\
    CUHK 
 	\and
 	Gauri Joshi\footnotemark[3] \\
	CMU 
	\and
 	Yuejie Chi\thanks{Department of Statistics and Data Science, Yale University.} \\
	Yale
}

\date{\today}

\maketitle

\input{./abstract}

\smallskip


\setcounter{tocdepth}{2}
\tableofcontents

\input{./intro.tex}

\input{./formulation.tex}

\input{./single-agent.tex}

\input{./federated.tex}

\input{./discussion.tex}

\section*{Acknowledgements}

The work of Y. Chi is supported in part by NSF under grants ECCS-2537189 and CNS-2148212, and in part by funds from federal agency and industry partners as specified in the Resilient \& Intelligent NextG Systems (RINGS) program. The work of G. Joshi is supported in part by the grants NSF-CCF 2007834, CCF-2045694, CNS-2112471, and ONR N00014-23-1-2149.
The work of G. Li is supported in part by the Chinese University of Hong Kong Direct Grant for Research and the Hong Kong Research Grants Council ECS 2191363 and GRF 2131005.

\bibliography{bibfileRL}
\bibliographystyle{apalike} 

\appendix

\input{./analysis.tex}

\input{./analysis_fl.tex}
\input{./analysis_policy.tex}

\end{document}

%% file: abstract.tex
\begin{abstract}
        Average-reward reinforcement learning offers a principled framework for long-term decision-making by maximizing the mean reward per time step. Although Q-learning is a widely used model-free algorithm with established sample complexity in discounted and finite-horizon Markov decision processes (MDPs), its theoretical guarantees for average-reward settings remain limited. This work studies a simple but effective Q-learning algorithm for average-reward MDPs with finite state and action spaces under the weakly communicating assumption, covering both single-agent and federated scenarios. For the single-agent case, we show that Q-learning with carefully chosen parameters achieves sample complexity $\widetilde{O}\left(\frac{|\mathcal{S}||\mathcal{A}|\|h^{\star}\|_{\mathsf{sp}}^3}{\varepsilon^3}\right)$, where $\|h^{\star}\|_{\mathsf{sp}}$ is the span norm of the bias function, improving previous results by at least a factor of $\frac{\|h^{\star}\|_{\mathsf{sp}}^2}{\varepsilon^2}$. In the federated setting with $M$ agents, we prove that collaboration reduces the per-agent sample complexity to $\widetilde{O}\left(\frac{|\mathcal{S}||\mathcal{A}|\|h^{\star}\|_{\mathsf{sp}}^3}{M\varepsilon^3}\right)$, with only $\widetilde{O}\left(\frac{\|h^{\star}\|_{\mathsf{sp}}}{\varepsilon}\right)$ communication rounds required.
        These results establish the first federated Q-learning algorithm for average-reward MDPs, with provable efficiency in both sample and communication complexity.
\end{abstract}

%% file: intro.tex
\section{Introduction}

Reinforcement learning (RL) seeks to maximize cumulative rewards through sequential decision-making. While most RL research focuses on the discounted reward framework, many practical applications require optimizing long-run average reward, which better reflects sustained performance rather than short-term gains. Average-reward RL \citep{howard1960dynamic} does not rely on an arbitrary discount factor and instead focuses on the agent's steady-state behavior over an infinite horizon, which introduces unique algorithmic and theoretical challenges. Recent work has begun to address these, with sample efficiency and feasibility remaining key concerns. 
 
Most existing average-reward RL algorithms are model-based \citep{jin2021towards,wang2022near,wang2023optimal,zurek2024span,zurek2024plug,zurek2025span,tuynman2024finding} or require prior knowledge of environment-specific parameters, such as mixing time or span norm \citep{jin2020efficiently,zhang2023sharper,li2024stochastic}, which are typically unknown and hard to estimate. Model-free methods avoid such requirements, but existing guarantees often suffer from suboptimal sample complexity \citep{jin2024feasible,bravo2024stochastic} or require sophisticated algorithmic designs, such as variance reduction techniques \citep{lee2025near}. While Q-learning \citep{watkins1989learning} stands out for its simplicity and memory efficiency and has proven effective in the discounted setting \citep{li2024q}, its sample efficiency and theoretical guarantees for average-reward RL remain limited.
 
One promising direction to improve sample efficiency is the federated framework \citep{kairouz2021advances}, where multiple agents collaboratively learn value functions and policies using distributed data and computation. While federated RL has shown significant sample efficiency gains in the discounted reward settings \citep{khodadadian22fedrlspeedup,woo2023blessing,salgia2024sample,zheng2024federated,zhang2024finite}, its extension to average-reward RL is particularly compelling since the inherently longer effective horizon in average-reward problems amplifies sample requirements, making federated learning a practical and scalable solution to address these challenges without introducing additional algorithmic complexity.

In this work, we propose a simple yet effective federated Q-learning algorithm tailored for average-reward RL in the synchronous setting with a generative model. Our method improves sample efficiency through principled scheduling of learning rates, discount factors, and communication intervals, while maintaining low communication cost.
To the best of our knowledge, this is the first work to develop federated Q-learning algorithms for average-reward MDPs, achieving improved sample and communication efficiency. Our results provide a first step toward addressing the sample efficiency challenges of average-reward RL via collaboration among multiple agents in a federated manner.

\subsection{Our contribution}
We introduce a simple yet effective Q-learning algorithm for average-reward MDPs, applicable to both single-agent and federated settings. %
Formally, we consider tabular average-reward MDPs with state space $\mathcal{S}$, action space $\mathcal{A}$, and a finite span norm of the bias function $\|h^{\star}\|_{\mathsf{sp}}$. There are $M \ge 1$ agents, each equipped with a generative model and synchronous updates, collaborating in a federated manner to learn the $\varepsilon$-optimal average reward $J^{\star}$ and policy $\pi^{\star}$.
Our main contributions are summarized as follows.
\begin{itemize}[leftmargin=*]
\item \textbf{Single-agent average-reward Q-learning with tighter sample complexity.}
We introduce two distinct parameter choices for single-agent Q-learning tailored to average-reward MDPs, each offering flexible scheduling of learning rates and discount factors.
Without any algorithmic changes to vanilla Q-learning, we show that principled choices of the discount factor and learning rate schedules alone yield improved sample complexity over existing Q-learning algorithms. Specifically, our Q-learning algorithm attains an $\varepsilon$-approximation to the optimal average reward $J^{\star}$ with sample complexity (up to logarithmic factors)
$$
\widetilde{O}\left(\frac{|\mathcal{S}||\mathcal{A}|\|h^{\star}\|_{\mathsf{sp}}^3}{\varepsilon^3}\right),
$$
exhibiting an improvement over the sample complexity of \citet{jin2024feasible} by a factor of $\widetilde{O}(\|h^{\star}\|_{\mathsf{sp}}^2/\varepsilon^{2})$.

\item  \textbf{Federated average-reward Q-learning with linear speedup.}
We propose a federated variant of average-reward Q-learning algorithm, which collaboratively learns Q-values across $M$ agents via independent local updates at agents and periodic global aggregation at a central server with judiciously designed learning rates and discount factors for the federated setting. We establish that this federated approach achieves an $\varepsilon$-approximation to the optimal average reward $J^{\star}$ with sample complexity per agent (up to logarithmic factors)
\begin{align} \label{eq:sample-complexity-fl}
\widetilde{O}\left(\frac{|\mathcal{S}||\mathcal{A}|\|h^{\star}\|_{\mathsf{sp}}^3}{M\varepsilon^3}\right),
\end{align}
demonstrating linear speedup with respect to the number of agents $M$. %

\item \textbf{Efficient communication scheduling for federated average-reward Q-learning.}
We present two communication scheduling strategies corresponding to the two parameter regimes in our federated Q-learning algorithm, each minimizing communication rounds while maintaining the convergence speedup. We show that the number of communication rounds required to achieve $\varepsilon$-optimal values, with per-agent sample complexity as in \eqref{eq:sample-complexity-fl}, is (up to logarithmic factors) bounded by
$$
\widetilde{O}\left(\frac{\|h^{\star}\|_{\mathsf{sp}}}{\varepsilon}\right).
$$
This communication complexity is independent of the number of agents $M$, indicating that increasing the number of agents does not lead to additional communication overhead, thereby ensuring scalability. 
Moreover, our analysis shows that both communication scheduling strategies achieve the same communication complexity, offering flexibility for practical deployment.

\end{itemize}

\subsection{Related work}

\paragraph{Average-reward MDPs.}
The average-reward MDP framework was first introduced by \citet{howard1960dynamic}, and it has been investigated in various settings.
Model-based algorithms for average-reward MDPs include \citet{jin2021towards,wang2022near,wang2023optimal,zurek2024span,zurek2024plug,zurek2025span,tuynman2024finding}, while model-free methods are studied by \citet{ganesh2024sharper,wei2020model,wan2021learning,wan2024convergence,chen2025non}. Policy gradient approaches are explored in \citet{bai2024regret,kumar2024global}. Sample complexity lower bounds are established by \citet{jin2021towards,wang2022near,bravo2024stochastic}. Most prior work focuses on the synchronous setting with a generative model, though some, such as \citet{chen2025non}, consider the asynchronous setting with Markovian trajectories.
Many existing methods rely on prior knowledge of problem-dependent parameters, such as the mixing time or the span of the bias function, which are typically challenging to estimate in practice.
Recently, \citet{lee2025near} developed a model-free algorithm that achieves near-optimal sample complexity for average-reward MDPs without requiring prior knowledge of problem parameters, leveraging recursive sampling as a variance reduction technique. 
While effective, this approach introduces additional algorithmic and computational complexity, which may limit its practical adoption. In contrast, \citet{jin2024feasible} proposed a simpler adaptation of vanilla Q-learning, relying only on dynamic updates of horizon factors and learning rates. However, their method remains suboptimal in terms of sample efficiency.

\paragraph{Sample complexity of Q-learning for discounted MDPs.} Q-learning \citep{watkins1992q} is one of the most widely studied model-free RL algorithms, particularly in the context of discounted MDPs. The finite-time sample complexity of Q-learning has been analyzed in various settings: \citet{even2003learning,beck2012error,li2024q,wainwright2019stochastic} have investigated the synchronous case with a generative model, \citet{szepesvari1998asymptotic,li2021sample,yan2022efficacy,qu2020finite,shi2022pessimistic} have focused on asynchronous and offline Q-learning, while \citet{jin2018q} studied online Q-learning, to name only a few. In addition, \citet{wainwright2019variance,sidford2018variance,li2021sample,shi2022pessimistic,li2021breaking,zhang2020almost} have proposed variance-reduced variants of Q-learning and improved sample complexity bounds.

\paragraph{Federated RL.}
Federated RL enables multiple agents to collaboratively learn optimal policies, addressing limitations in sample availability and computational resources. Recent work has studied sample complexity improvements in discounted and finite-horizon MDPs, showing that federated Q-learning achieves linear speedup with respect to the number of agents, without data sharing \citep{khodadadian22fedrlspeedup,woo2025blessing,salgia2024sample}. In addition, \citet{zheng2024federated} analyzed regret for online federated Q-learning, \citet{woo2024federated} studied offline variants and the impact of collective data coverage, \citet{yang2024federated,zhang2024finite} has explored federated RL algorithms in heterogeneous environments. In particular, \citet{naskar2025parameter} proposed a federated TD learning algorithm for average-reward settings in heterogeneous environments.
Also, communication efficiency is also crucial; \citet{salgia2024sample} established lower bounds on communication complexity for discounted MDPs.

\begin{table*}[!t]
\centering
\resizebox{\textwidth}{!}{
\begin{tabular}{|ccccc|} 
 \hline
{\bf Previous works} &{\bf Sample complexity}& {\bf Number of agents }&{\bf MDP class} & {\bf Target error} \\ 
 \hline\hline&&&&\\[-2.1ex]
 \citet{jin2021towards}   & $\widetilde{\Omega}\left({|\cS||\cA|t_{\mathsf{mix}}}/{\varepsilon^{2}}\right)$ & $1$ & U & $\|J^{\hat{\pi}}-J^{\star}\|_{\infty}$ \\[0.1ex]
 \citet{wang2022near}  & $\widetilde{\Omega}\left({|\cS||\cA|\sp{h^\star}}/{\varepsilon^{2}}\right)$ & $1$  & W & $\|J^{\hat{\pi}}-J^{\star}\|_{\infty}$ \\ \hline\hline&&&&\\[-2.1ex]
 \citet{bravo2024stochastic}& $\widetilde{O}\left({|\cS||\cA|\sp{h^\star}^7}/{\varepsilon^{7}}\right)$ & $1$  &W & $\|J^{\hat{\pi}}-J^{\star}\|_{\infty}$ \\[0.1ex]
\citet{lee2025near}& $\widetilde{O}\left({|\cS||\cA|\sp{h^\star}^{2}}/{\varepsilon^{2}}\right)$ & $1$ &W & $\|J^{\hat{\pi}}-J^{\star}\|_{\infty}$ \\[0.1ex]
\citet{jin2024feasible}& $\widetilde{O}\left({|\cS||\cA|t^8_{\mathsf{mix}}}/{\varepsilon^{8}}\right)$ & $1$ & U & $\|J^{\hat{\pi}}-J^{\star}\|_{\infty}$ \\[0.1ex]
Ours (Theorem~\ref{thm:avg-policy-fed})& $\widetilde{O}\left({|\cS||\cA|\sp{h^\star}^5}/{M\varepsilon^{5}}\right)$ & $M$ &W & $\|J^{\hat{\pi}}-J^{\star}\|_{\infty}$ \\[0.6ex]
\hline\hline
&&&&\\[-2.1ex]
\citet{jin2024feasible}& $\widetilde{O}\left({|\cS||\cA|t^5_{\mathsf{mix}}}/{\varepsilon^{5}}\right)$ & $1$ &U  & $\|\hat{Q}-J^{\star}\|_{\infty}$ \\[0.1ex]
Ours (Theorem~\ref{thm:average-reward}) & $\widetilde{O}\left({|\cS||\cA||\sp{h^\star}^3}/{\varepsilon^{3}}\right)$ & $1$ &W  & $\|\hat{Q}-J^{\star}\|_{\infty}$ \\[0.1ex]
Ours (Theorem~\ref{thm:average-reward-fed})& $\widetilde{O}\left({|\cS||\cA|\sp{h^\star}^3}/{M\varepsilon^{3}}\right)$ & $M$ &W & $\|\hat{Q}-J^{\star}\|_{\infty}$  \\[0.6ex]
  \hline
\end{tabular}
}
\caption{{\bf Algorithms for average-reward MDPs.} The table summarizes the leading-order sample complexity of model-free algorithms for obtaining an $\varepsilon$-optimal value or policy with probability at least $1-\delta$ in the synchronous setting without prior knowledge, where $\widetilde{O}(\cdot)$ omits logarithmic factors. The `MDP class' column indicates the applicable MDP type: (U) uniformly mixing or ergodic MDPs with finite mixing time $t_{\mathsf{mix}}$; (W) weakly communicating MDPs with bias vector $h^\star$. }
\label{tab:comparison}
\end{table*}

\paragraph{Notation.}
For a finite set $\mathcal{S}$, let $\Delta(\mathcal{S})$ denote the probability simplex over $\mathcal{S}$. For any positive integer $I > 0$, define $[I] \coloneqq \{1, \ldots, I\}$. For $x \in \mathbb{R}^I$, the infinity norm is given by $\|x\|_{\infty} = \max_{i \in I} |x(i)|$, and the span seminorm by $\sp{x} = \max_{i \in I} x(i) - \min_{i \in I} x(i)$. We write $f(\cdot) = \widetilde{O}(g(\cdot))$ or $f \lesssim g$ (respectively, $f(\cdot) = \widetilde{\Omega}(g(\cdot))$ or $f \gtrsim g$) to indicate that $f(\cdot)$ is, up to logarithmic factors, at most (respectively, at least) order $g(\cdot)$. The notation $f \asymp g$ means that both $f \lesssim g$ and $f \gtrsim g$ hold.

%% file: formulation.tex
\section{Problem Formulation}
\label{sec:formulation}

\subsection{Average-reward Markov Decision Process}
An infinite-horizon Average-reward Markov Decision Process (AMDP) is represented by a tuple $\mathcal{M} = \{\mathcal{S},\mathcal{A},P,r\}$, where $\mathcal{S} = \{1,\ldots,S\}$ and $\mathcal{A} = \{1,\ldots,A\}$ denote the finite state and  action spaces, respectively.
The transition kernel $P:\mathcal{S}\times\mathcal{A}\mapsto \Delta(\mathcal{S})$ specifies the probability distribution over the next state given a state-action pair, i.e., $P(\cdot|s,a)\in\Delta(\mathcal{S})$ denotes the transition probability when action $a$ is taken in state $s$.
The reward function $r:\mathcal{S}\times\mathcal{A}\mapsto[0,1]$ assigns a deterministic and bounded immediate reward $r(s,a)$ to each state-action pair $(s,a)\in\mathcal{S}\times\mathcal{A}$.

\paragraph{Average reward and bias function.} A deterministic policy is a mapping $\pi:\mathcal{S}\mapsto\mathcal{A}$, where $\pi(s)$ denotes the action taken in state $s$.
Under policy $\pi$, the long-term average reward starting from an initial state $s\in\mathcal{S}$ is defined as
\begin{align*}
J^{\pi}(s) \coloneqq \liminf_{T\to \infty}\mathbb{E}_{\pi}\left[\frac1T\sum_{t=0}^{T-1} r(s_t,a_t)|s_0=s\right],\quad \forall s\in\mathcal{S},
\end{align*}
where the trajectory $\{s_t,a_t\}_{t=0}^{\infty}$ is generated by following policy $\pi$ and evolving according to the transition kernel $P$.
Another quantity of interest in AMDP is the bias function of policy $\pi$, defined as
$$
h^{\pi}(s) \coloneqq \text{C-}\hspace{-0.3em}\lim_{T\to\infty}\mathbb{E}_{\pi}\left[\sum^{T-1}_{t=0} (r(s_t,a_t) -J^{\pi}(s_t))\right],
$$
where C-$\lim$ denotes the Cesaro limit. 
If the Markov chain induced by $\pi$ is aperiodic, then C-$\lim$ coincides with the usual limit.

The objective of average-reward RL is to find an optimal policy $\pi^{\star}$ that maximizes the average reward:
\begin{align*}
\pi^{\star} = \arg\max_{\pi} J^{\pi}.
\end{align*}
For convenience, we denote the optimal value by $J^{\star} = J^{\pi^{\star}}$.
The corresponding bias function under policy $\pi^{\star}$ is denoted as $h^{\star}$. 

\paragraph{Weakly communicating MDPs.} A weakly communicating MDP is one in which there exists a set of states such that, under some policy, each state in the set is reachable from every other state in that set, and states outside this set are transient under all policies \citep{puterman2014markov}.
For weakly communicating MDPs, the $\lim\inf$
in the definition of 
$J^{\pi}$ can be replaced by the usual limit.
Moreover, there always exists a unichain optimal policy, such that the optimal average reward $J^{\star}(s)$ are identical across all states $s\in\mathcal{S}$ \citep{puterman2014markov,zurek2024span}.
In this setting, it is well-known that $h^{\star}$ satisfies the following Bellman operator:
\begin{align}\label{eq-bellman}
J^{\star}+h^{\star}(s) = \max_{a\in\mathcal{A}}\left[r(s,a) + \sum_{s'\in\mathcal{S}}P(s'|s,a)h^{\star}(s')\right],
\end{align}
where we omit the state argument of $J^{\star}$ due to its constancy, and treat $J^{\star}$ as a scalar or a vector with dimension implied by the context throughout the remainder of the paper.
The span norm of bias function $h^{\star}$ is often used to characterize the sample complexity of reinforcement learning algorithms:
$$
\|h^{\star}\|_{\mathsf{sp}}\coloneqq \max_s h^{\star}(s) - \min_s h^{\star}(s).
$$

\paragraph{Mixing time.} 
For comparison purpose, we briefly recall the notion of mixing time, a stronger assumption than weakly communicating MDPs and widely used in prior literature.
Intuitively, the mixing time $t_{\mathsf{mix}}$ measures how quickly the Markov chain induced by any policy converges to its stationary distribution. 
This requirement is stronger than the weakly communicating assumption, since it implies uniform ergodicity of the chain.
In particular, it has been shown that \citep{wang2022near} 
$$
\|h^{\star}\|_{\mathsf{sp}}\lesssim t_{\mathsf{mix}}.
$$
In this paper, we only assume weakly communicating MDPs.

\subsection{Discounted MDPs}
Many prior works \citep{jin2021towards,jin2024feasible} on analyzing AMDPs build upon well-established techniques developed for Discounted Markov Decision Processes (DMDPs), where future rewards are geometrically weighted by a discount factor to emphasize recent rewards. A discounted MDP is defined by the tuple $(\mathcal{S}, \mathcal{A}, P, r, \gamma)$, where $\gamma \in [0,1)$ denotes the discount factor.
Given a discount factor $\gamma$ and a policy $\pi$, the \emph{discounted} value function $V_{\gamma}^{\pi}: \mathcal{S} \mapsto \mathbb{R}$ and the discounted Q-function $Q_{\gamma}^{\pi}: \mathcal{S} \times \mathcal{A} \rightarrow \mathbb{R}$ represent the expected cumulative discounted rewards starting from a state $s$ and from a state-action pair $(s,a)$, respectively:
\begin{align} \label{eq:def_V}
   V_{\gamma}^{\pi}(s) &\defn (1-\gamma) \mathbb{E} \left[ \sum_{t=0}^{\infty} \gamma^t r(s_t,a_t ) \,\big|\, s_0 =s \right], \\
  Q_{\gamma}^{\pi}(s,a) &\defn (1-\gamma)  \mathbb{E} \left[  \sum_{t=0}^{\infty} \gamma^t r(s_t,a_t ) \,\big|\, s_0 =s, a_0 = a \right].
\end{align}
When the rewards lie within $[0,1]$, it follows that for any policy $\pi$, 
\begin{equation*}
 0\leq V_{\gamma}^{\pi} \leq 1  , \qquad 0 \leq Q_{\gamma}^{\pi} \leq 1.
\end{equation*}
We note that the above definition differs from their typical forms \citep{li2021tightening} by a factor of $1/(1-\gamma)$. 
The optimal policy for the $\gamma$-discounted reward setting is defined as the policy that maximizes the discounted value function across all states, denoted by $\pi_{\gamma}^{\star}$. The existence of such an optimal policy is guaranteed \citep{puterman2014markov}, and this policy simultaneously maximizes the Q-function. The corresponding optimal value function and Q-function are denoted by $V_{\gamma}^{\star} := V_{\gamma}^{\pi_{\gamma}^{\star}}$ and $Q_{\gamma}^{\star} := Q_{\gamma}^{\pi_{\gamma}^{\star}}$, respectively.
The discounted setting can be connected to the average-reward setting by taking the limit as $\gamma \to 1$, where each entry of $Q_{\gamma}^\star$ converges to the optimal average reward $J^\star$. This connection is closely related to the notion of Blackwell optimality \citep{puterman2014markov}.

%% file: single-agent.tex
\section{Single-agent Average-reward Q-Learning}
\label{sec:single}

In this section, we first provide a brief introduction of the sampling scheme and the Q-learning algorithm considered in this study, and subsequently present the corresponding sample complexity analysis.

\subsection{Synchronous sampling with a generative model}

In this paper, we focus on the synchronous sampling, where all state-action pairs are sampled uniformly assuming access to a generative model or a simulator \citep{kearns1999finite}. In every iteration $t$, an agent generates a transition sample
 \begin{equation}\label{eq:sync_sampling}  
s_t(s,a) \sim P(\cdot|s,a),\quad \forall (s,a) \in \cS \times \cA.
 \end{equation}
 
\paragraph{Synchronous Q-learning for average-reward RL.} Q-learning \citep{watkins1992q} stands out as one of the most widely used model-free approaches, which directly estimates the Q-function without needing model estimation or prior knowledge. In this paper, we focus on a Q-learning approach to directly estimate the optimal value $J^{\star}$ for average-reward RL. 
In the synchronous setting, Q-learning operates as follows: beginning with an initial Q-function $Q_0$, the algorithm updates the Q-function at each iteration $t \geq 1$ using the following rule:
\begin{align} \label{eq:sync-q-learning}
\forall (s,a)\in \mathcal{S} \times \mathcal{A} : \qquad 
Q_{t}(s, a) &= (1-\eta_t) Q_{t-1}(s,a) + \eta_t \bigg( (1-\gamma_t) r(s,a) 
 + \gamma_t \max_{a' \in \mathcal{A}} Q_{t-1}(s_t(s,a),a') \bigg),
\end{align}
where $s_t(s,a)\sim P(\cdot|s,a)$ represents an independent sample drawn for each state-action pair $(s,a) \in \mathcal{S} \times \mathcal{A}$, while $\eta_t$ and $\gamma_t$ correspond to the learning rate and discount factor at iteration $t$, respectively. A key distinction from traditional Q-learning in discounted MDPs---which applies a constant discount factor to future values---is that Q-learning for average-reward RL necessitates dynamic adjustment of discount factors when evaluating future values \citep{jin2024feasible}. 
 
 \subsection{Algorithm}

We consider a stage-wise Q-learning algorithm in which the discount factor varies dynamically across epochs. 
To be specific, in the
$t$-th iteration of the $k$-th epoch, we maintain an estimate $Q_{k,t} \in\mathbb{R}^ {S\times A}$ 
and update it using a variant of the standard Q-learning update rule, with an epoch-dependent discount factor $\gamma_k$, sample size $N_k$, and the step-size $\eta_{k,t}$:
\begin{align} \label{eq:single-q-learning}
\forall (s,a) \in \cS \times \cA: \qquad 
Q_{k,t}(s, a) &= (1-\eta_{k,t}) Q_{k,t-1}(s,a) + \eta_{k,t} \bigg( (1-\gamma_k) r(s,a)
 + \gamma_k V_{k,\iota(k,t)}(s_{k,t}(s,a)) \bigg),
\end{align}
where $s_{k,t}(s,a)\sim P(\cdot|s,a)$.
 The value function $V_{\iota(k,t)}(s_{k,t}(s,a))\coloneqq\max_{a' \in \mathcal{A}} Q_{\iota(k,t)}(s_{k,t}(s,a),a')$ used in the update is calculated using a historical estimate $Q_{\iota(k,t)}$, where $\iota(k,t)$ denotes a historical index.

\begin{algorithm}[t]
\renewcommand{\algorithmicrequire}{\textbf{Input:}}
\renewcommand{\algorithmicensure}{\textbf{Output:}}
\caption{Average-reward Q-learning}
\label{alg:average-reward}
\begin{algorithmic}[1]
\REQUIRE number of epochs $K$, discount factor $\{\gamma_k\}_{k=1}^K$, sample size $\{N_k\}_{k=0}^K$, step-size $\{\eta_{k,t}\}_{(k,t)=(1,1)}^{(K,N_k)}$, and historical index $\iota(k,t)$.
\ENSURE average reward estimation $Q_K$.
\STATE Initialize $Q_{0,0}=0$, $V_{0,0}=0$.\\
\FOR{$k=1,\cdots,K$}
\STATE Initialize $Q_{k,0} = Q_{k-1,N_{k-1}}$.
\FOR{$t=1,\cdots,N_k$}
\STATE Draw $s_{k,t}(s,a)\sim {P}(\cdot|s,a)$,~$\forall (s,a)\in\mathcal{S}\times\mathcal{A}$.
\STATE Compute $Q_{k,t}(s,a)$ according to \eqref{eq:single-q-learning}.
\STATE Compute $V_{k,t}(s) = \max_{a'\in\mathcal{A}}Q_{k,t}(s,a')$ for all $s\in\mathcal{S}$.
\ENDFOR
\ENDFOR
\end{algorithmic}
\end{algorithm}

It is obvious that the choice of the parameters--- including $\gamma_k$, $N_k$, $\eta_{k,t}$ and $\iota(k,t)$--- has a significant influence on the algorithm's performance.
In the following, we shall provide two groups of parameters that do not require prior knowledge of the mixing time or $\|h^{\star}\|_{\mathsf{sp}}$.
Assume $N_0 = 0$. For some sufficiently large constant $c_N>0$, two groups of parameters are as follows.
\begin{itemize}[leftmargin=*]
\item 
The first group decays the discount factor $1-\gamma_k$ at an exponential rate, such that the initial estimate $Q_{k,0}$ converges to the average reward $J^{\star}$ exponential fast.
In each epoch, the step-size $\eta_{k,t}$ and the historical index $\iota(k,t)$ are chosen in a manner similar to those in the standard Q-learning algorithm, such that $Q_{k,t}$ converges to the $\gamma_k$-discounted optimal Q-function $Q_{\gamma_k}^{\star}$at a comparable rate. The sample size $N_k$ is selected to ensure that $Q_{k,N_k}$ sufficiently approximates $Q_{\gamma_k}^{\star}$.
\begin{subequations}\label{eq:para-single}
\begin{align}\label{eq:para-group-1}
N_k &= c_N 2^k, \quad \gamma_k = 1-\frac{2\log(4N_k)}{N_k^{1/3}}, \notag \\
\eta_{k,t} & = \frac{1}{1+\frac{t^{2/3}}{8\log (4t)}},\quad \iota(k,t) = (k,t-1). 
\end{align}
\item The second group decays the discount factor $1-\gamma_k$ at a rate of $1/k$, such that the difference between $Q_{k,0}$ and $J^{\star}$ decreases at the same rate.
In each epoch, the step-size $\eta_{k,t}$ and the historical index $\iota(k,t)$ are fixed, resulting in a convergence rate of $1/\sqrt{N_k}$ for $Q_{k,t}$.
Thus the sample size $N_k$ is chosen to be in the order of $k^2$:
\begin{align}\label{eq:para-group-2}
N_k &= c_Nk^2\log^5(k+1)\log^3\left(\frac{SA}{\delta}\right),\quad 
\gamma_k = \frac{k}{k+1},\notag \\
\eta_{k,t} &= \frac{1}{1+\frac{N_k}{4\log (\sum_{i=1}^kN_i)}},\quad \iota(k,t) = (k-1,N_{k-1}).
\end{align}
\end{subequations}
\end{itemize}

\subsection{Analysis: optimal value estimation}

The following theorem presents the sample complexity of Algorithm \ref{alg:average-reward}, under the mild assumption of a weakly communicating MDP.
The proof is provided in Appendix \ref{sec:proof-single}.

\begin{theorem}\label{thm:average-reward}
Assume $\|h^{\star}\|_{\mathsf{sp}} < \infty$. For both of the two groups of parameters in \eqref{eq:para-single}, 
with probability at least $1-\delta$, the output of Algorithm \ref{alg:average-reward} satisfies
\begin{align*}
\left\|Q_{K,N_K} - J^{\star}\right\|_{\infty}\le \frac{ C\|h^{\star}\|_{\mathsf{sp}}}{(T_K)^{1/3}}
\log^4{\frac{4SAT_k }{\delta}},
\end{align*}
where $C$ is a positive constant, $T_K\coloneqq \sum_{k=1}^K N_k$ denotes the total number of iterations.
For $\varepsilon \in (0,1]$, to achieve $\|Q_{K,N_K} - J^{\star}\|_{\infty}  \le \varepsilon$, the total number of samples is bounded as
$$
SAT_K = \widetilde{O}\left(SA \frac{ \|h^{\star}\|_{\mathsf{sp}}^3}{\varepsilon^3}\right).
$$
\end{theorem}
 
Theorem \ref{thm:average-reward} shows that the proposed Q-learning algorithm attains a sample complexity of $\widetilde{O}(SA\|h^{\star}\|_{\mathsf{sp}}^3/\varepsilon^3)$ for both parameter groups. This is reminiscent to the discounted setting \citep{li2024q}, where different learning rate schedules, such as constant or rescaled linear rates, achieve the same optimal rate.
Moreover, this result provides the best known sample complexity for Q-learning algorithms in average-reward RL, improving upon the bound in \citet{jin2024feasible} by a factor of $\widetilde{O}(\|h^{\star}\|_{\mathsf{sp}}^2/\varepsilon^{2})$.

Previous studies have established stronger results by leveraging more sophisticated algorithmic techniques.
For example, \citet{jin2021towards} extended learning algorithms with optimal sample complexity in discounted MDPs to the average-reward setting, thereby achieving better bounds.
\citet{zhang2023sharper} refined Q-learning via the use of the upper confidence bound (UCB) technique and obtained an improved dependency on $\|h^{\star}\|_{\mathsf{sp}}$ and $\varepsilon$.
Policy optimization methods adopted by \citet{wang2017primal, li2024stochastic} achieved a $\widetilde{O}(1/\varepsilon^2)$ rate.
All of them required prior knowledge of the mixing time or $\|h^{\star}\|_{\mathsf{sp}}$.
Moreover, 
\citet{lee2025near} established the state-of-the-art by incorporating the Halpern iteration with variance reduction, implemented via differential techniques with sample batching. 

In contrast, our result focuses on the vanilla Q-learning algorithm, without variance reduction or UCB enhancements. The performance is constrained by the behavior of Q-learning in discounted MDPs, which is known to be suboptimal \citep{li2024q}, explaining why our bound falls short of the optimal rate. Nevertheless, Q-learning remains a classical and fundamental algorithm, valued for its simplicity of implementation and practical applicability. 
A rigorous theoretical analysis of its behavior is therefore essential, and it provides foundation for subsequent analysis with variance reduction or UCB techniques.

%% file: federated.tex
\section{Federated Average-reward Q-Learning}
\label{sec:federated}

Average-reward RL presents unique computational challenges because the optimal value function and policy depend on long-term average performance, which requires substantially more samples and training iterations to achieve optimality and places a significant burden on a single agent. Federated learning addresses these challenges by distributing the computational and sampling load across multiple agents, enabling collaborative learning of a shared model without directly sharing local datasets at agents.

In this section, we develop federated variants of average-reward Q-learning involving $M$ agents and analyze the sample complexity of the variants. We study a federated framework where $M$ agents independently update their local Q-function estimates using their respective datasets or generative models, while periodically communicating with a central server to aggregate these local estimates into a global model. Through this collaborative approach, our objective is to efficiently learn the optimal average reward $J^{\star}$ or optimal policy $\pi^{\star}$ by leveraging the combined learning experience of all participating agents.

\subsection{Algorithm}
We first begin with the federated variants of Q-learning in the synchronous setting with generative models, where all the state-action pairs are updated simultaneously at all agents. In the synchronous setting, at every epoch $k \in [K]$ and iteration $t \in [N_k]$, each agent $ m \in [M]$ has access to a generative model, and generates a new sample
 \begin{equation}\label{eq:sync_sampling-fed}  
s_{k,t}^m(s,a) \sim P(\cdot|s,a)
\end{equation}
for every state-action pair $(s,a) \in \cS \times \cA$ independently. We assume agents communicate their estimates to the central server only at iterations when communication is scheduled, and we denote the communication schedule at epoch $k$ as $\synset(k)$. Then, synchronous federated Q-learning proceeds according to the following steps.
\begin{enumerate}[leftmargin=12pt] 
\item {\em Local updates:} For each epoch $k$, local Q-functions at agents are initialized as $Q_{k,0}^m = Q_{k-1}$, where $Q_{k-1}$ is a global Q-estimate computed in the previous epoch. Then, for each iteration $t \in [N_k]$, each agent independently updates {\em all} entries of its Q-estimate $Q_{k,t-1}^m$ to reach some {\em intermediate} estimate following the update rule:
\begin{align} \label{eq:syncq-local}
\forall (s,a) \in \cS \times \cA: \quad  
Q_{k,t}^m(s, a) &= (1-\eta_{k,t}) Q_{k,t-1}^m(s,a) + \eta_{k,t} \bigg( (1-\gamma_{k}) r(s,a)
 + \gamma_{k}   V_{k,\syn(k,t)}(s_{k,t}^m(s,a) )  \bigg) ,  
\end{align}
where $s_t^m(s,a)$ is drawn according to \eqref{eq:sync_sampling-fed}, and $\eta_{k,t}$ and $\gamma_{k}$ are the learning rate and discount rate scheduled for epoch $k$ and iteration $t$ at all agents $m\in[M]$. Here, $\syn(k,t)$ is the most recent iteration before $t$ when agents communicate with the central server, where $\syn(k,t) = \max\{j \in \synset(k) : j < t\}$ if such $j$ exists, and $\syn(k,t) = 0$ otherwise.

\item {\em Global aggregation:} At each stage $k$, whenever $t \in \synset(k)$ and communication is scheduled, all agents send their local Q-function estimates to the central server. The server then aggregates these by averaging the local Q-estimates $Q_{k,t}^m$ from all agents to form the updated global estimate $Q_{k,t}$:
  \begin{align} \label{eq:syncq-periodic}
\forall (s,a): \quad  Q_{k,t}(s,a) =
    \frac{1}{M} \sum_{m=1}^M Q_{k,t}^m(s,a).
\end{align}
\end{enumerate}
Accordingly, after aggregation, the global value function is updated as $V_{k,t}(s) = \max_{a}Q_{k,t}(s,a)$, and all agents synchronize their local Q-functions to this updated global Q-function. The complete procedure is summarized in Algorithm~\ref{alg:fed-average-reward}.

\begin{algorithm}[ht]
  \begin{algorithmic}[1] 
    \STATE \textbf{inputs:} number of agents $M$, number of epochs $K$, discount factor $\{\gamma_k\}_{k=1}^K$, sample size $\{N_k\}_{k=0}^K$, step-size $\{\eta_{k,t}\}_{(k,t)=(1,1)}^{(K,N_k)}$. 
    \STATE  Initialize $Q_{0,0}^m = 0$, $V_{0,0}^m = 0$~for~all~$m \in [M]$. 
    \FOR{$k=1,\cdots,K$}
    \STATE{Initialize $Q_{k,0}^m=Q_{k-1, N_{k-1}}$, $V_{k,0}^m=V_{k-1, N_{k-1}}$ for all $m \in [M]$}
    \FOR{$t=1,\cdots,N_k$}
    \FOR{$m \in [M]$}    
    \STATE{ Draw  $s_t^m(s,a) \sim  P(\cdot \mymid s,a)$,~$\forall (s,a)\in \cS\times \cA$. }
    \STATE{Compute $Q_{k,t}^m$ according to \eqref{eq:syncq-local}} 
    \STATE Compute $V_{k,t}^m(s) = \max_{a'\in\mathcal{A}}Q_{k,t}^m(s,a')$ for all $s\in \cS$
    \IF{$t \in \synset(k)$}
    \STATE{Compute global Q-estimates $Q_{k,t}$ via aggregation according to \eqref{eq:syncq-periodic}.}
    \STATE{Update global value estimates $V_{k,t}(s) = \max_{a}Q_{k,t}(s,a)$ for all $s \in \cS$.}
    \STATE{Synchronize local Q-estimates $Q_{k,t}^m = Q_{k,t}$ for all $m \in [M]$.}
    \ENDIF
    \ENDFOR
    \ENDFOR 
    \ENDFOR       
    \STATE \textbf{return:} ${Q}_{K, N_{K}}$. 
  \end{algorithmic} 
  \caption{Federated average-reward Q-learning}
  \label{alg:fed-average-reward}
\end{algorithm}

Assume $N_0 = 0$. In the federated setting with $M$ agents, for some sufficiently large constant $c_N>0$, two groups of parameters are given as below.
\begin{itemize}[leftmargin=*]
\item 
The first group of parameters accelerates the decay of the discount factor $(1-\gamma_k)$ by a factor of $M^{1/3}$, thereby increasing the effective planning horizon and expediting the convergence of $Q_{\gamma_k}^{\star}$ to $J^{\star}$. Simultaneously, the learning rate $\eta_{k,t}^m$ decays more gradually, also by a factor of $M^{1/3}$, reflecting the variance reduction achieved through collaborative learning among agents. These adjustments jointly reduce both the bias between $Q_{\gamma_k}^{\star}$ and $J^{\star}$ and the error in the convergence of the global Q-estimates $Q_{k,t}$ to $Q_{\gamma_k}^{\star}$, and crucially, balance these two sources of error so that they decrease at the same rate as $M$ increases, thereby improving overall convergence:
\begin{subequations}\label{eq:para-fed}
\begin{align}\label{eq:fed-para-group-1}
N_k &= c_N 2^k, \quad \gamma_k = 1-\frac{2\log(4MN_k)}{(MN_k)^{1/3}},\notag\\
\eta_{k,t} &= \frac{1}{1+\frac{t^{2/3}}{8M^{1/3}\log (4Mt)}},    \quad
\synset(k) = \{i \le N_k | i \equiv 0 ~(\text{mod}~ g_k) \} \cup \{N_k\},
\end{align}
where $g_k \defn \lceil \frac{-\log(1-\gamma_k)^2}{\eta_{k,N_k}} \rceil$.
\item
The second group retains the same rates for the discount factor $(1-\gamma_k)$ and learning rate $\eta_{k,t}^m$ as in the single-agent setting, but reduces the sample size $N_k$ by a factor of $M$. This reduction is due to the improved convergence rate of $Q_{k,t}$, as the variance reduction from multiple agents allows the algorithm to achieve the same error level with $M$ times fewer samples:
\begin{align}\label{eq:fed-para-group-2}
N_k &= c_N\max\left\{\frac{k^2}{M}\log^5(k+1)\log^3\left(\frac{SA}{\delta}\right),\, \log\left(Mk\log\frac{SA}{\delta} \right)\right\},\qquad
\gamma_k = \frac{k}{k+1},\notag\\
\eta_{k,t} &= \frac{1}{1+\frac{N_k}{4\log (M\sum_{i=1}^kN_i)}},\quad
\mathcal{C}_k =\{N_k\}.
\end{align}
\end{subequations}
\end{itemize}

\subsection{Analysis: optimal value estimation}
The next theorem provides a theoretical analysis of Algorithm \ref{alg:fed-average-reward}, characterizing both the sample complexity and the communication round requirements for estimating the optimal reward $J^{\star}$.

\begin{theorem}\label{thm:average-reward-fed}
Assume $\|h^{\star}\|_{\mathsf{sp}}<\infty$.
For both of the two groups of parameters in \eqref{eq:para-fed}, 
with probability at least $1-\delta$, the output of Algorithm \ref{alg:fed-average-reward} satisfies
\begin{align*}
\left\|Q_{K,N_K} - J^{\star}\right\|_{\infty}\le \frac{ C\|h^{\star}\|_{\mathsf{sp}}}{(MT_K)^{1/3}}\log^4{\frac{4MSAT_K }{\delta}},
\end{align*}
where $C$ is a positive constant, $T_K\coloneqq \sum_{k=1}^K N_k$ denotes the total number of iterations.
For $\varepsilon \in (0,1]$, to achieve $\|Q_{K,N_K} - J^{\star}\|_{\infty}  \le \varepsilon$, the total number of samples required per agent is bounded as
$$
SAT_K = \widetilde{O}\left( \frac{ SA\|h^{\star}\|_{\mathsf{sp}}^3}{M \varepsilon^3}\right),
$$
and the number of communication rounds can be bounded as
$$
  \sum_{k=1}^K |\synset(k)| = \widetilde{O}\left( \frac{\|h^{\star}\|_{\mathsf{sp}}}{\varepsilon} \right).$$

\end{theorem}

The proof of Theorem \ref{thm:average-reward-fed} is provided in Appendix \ref{sec:proof-federated}. We provide a few remarks on the implications as follows.

\paragraph{Sample complexity with linear speedup.}
Theorem \ref{thm:average-reward-fed} shows that federated collaboration among $M$ agents accelerates the convergence of Q-function estimates to the optimal average reward $J^{\star}$.
Specifically, compared to the single-agent case shown in Theorem~\ref{thm:average-reward}, the sample complexity per agent is reduced by a factor of $M$, demonstrating a linear speedup in convergence as the number of agents increases. 
Remarkably, this means that the federated algorithm achieves the same sample efficiency as if all data were sampled centrally, despite agents not sharing their individual datasets.

\paragraph{Communication efficiency.} 
In federated settings, communication between agents and the central server can be costly, making it crucial to minimize the communication frequency.
Theorem \ref{thm:average-reward-fed} indicates that the total number of communication rounds required to achieve an $\varepsilon$-optimal value is $\widetilde{O}(\|h^{\star}\|_{\mathsf{sp}}/\varepsilon)$, which is independent of the number of agents $M$. 
This means that adding more agents improves sample efficiency without increasing the frequency of communication rounds, allowing scalability in the number of agents without incurring additional communication costs.

Furthermore, the communication complexity bound holds for both parameter groups in \eqref{eq:para-fed}, enabling flexible choices of communication schedules based on resource constraints without compromising communication efficiency. For the first group, communication is scheduled periodically within each epoch at fixed intervals $g_k = \widetilde{O}\left( ( N_k^2/M)^{1/3}\right)$, with $N_k$ increasing exponentially. For the second group, communication occurs only at the end of each epoch, with the interval determined by the polynomially growing epoch size $N_k$. 
Depending on the available communication resources, communication rounds can be scheduled either at constant intervals with exponential jumps or at polynomially growing intervals.

\subsection{Analysis: optimal policy learning}
We now analyze the problem of learning an $\varepsilon$-optimal policy, providing sample complexity and communication complexity for learning a policy $\pi$ such that $J^{\star} - J^{\pi} \le \varepsilon$.

\begin{theorem} \label{thm:avg-policy-fed}
Assume $\|h^{\star}\|_{\mathsf{sp}}<\infty$.
Take $N_k = c_N 2^k$ for sufficiently large $c_N$, $\gamma_k = 1-\frac{1}{(N_kM)^{1/5}}$, $\eta_{k,t}=(1+\frac{t}{8(N_kM)^{1/5}\log(MN_k)})^{-1}$, and
$
\synset(k) = \left\{\left\lceil N_k\left(\frac{1+\gamma_k}{2}\right)^{i-1}\right\rceil
\bigg| i\in\left[\left\lceil\frac{4\log(1-\gamma_k)}{\log((1+\gamma_k)/2)}\right\rceil\right]\right\}$.
Take the output policy of Algorithm \ref{alg:fed-average-reward} as $\hat{\pi}(s) \coloneqq \arg\max_a Q_{K,N_K}(s,a)$.
Then with probability as least $1-\delta$, the average reward $J^{\hat{\pi}}$ satisfies
$
J^{\star} - J^{\hat{\pi}} \le \varepsilon
$
as long as 
$$
T_K = \sum_{k=1}^{K}N_k \gtrsim \frac{\|h^{\star}\|_{\mathsf{sp}}^5}{M\varepsilon^5}\log^5(N_KM)\log^{5/2}\frac{SAMT_K}{\delta}.
$$
The number of communication rounds is bounded as
$
\sum_{k=1}^K|\mathcal{C}_k| = \widetilde{O}\left(\|h^{\star}\|_{\mathsf{sp}}/\varepsilon\right).
$
\end{theorem}

Theorem \ref{thm:avg-policy-fed} shows that the algorithm can learn an $\varepsilon$-optimal policy with a sample complexity of $\widetilde{O}\left(\frac{SA\|h^{\star}\|_{\mathsf{sp}}^5}{M\varepsilon^5}\right)$ per agent, also achieving linear speedup in terms of the number of agents $M$. 
Notably, this improves the result of \citet{jin2024feasible} by a factor of $\widetilde{O}\big(\|h^{\star}\|_{\mathsf{sp}}^3/\varepsilon^{3}\big)$ in the single-agent case, where $M=1$.
The communication complexity remains $\widetilde{O}\left(\frac{\|h^{\star}\|_{\mathsf{sp}}}{\varepsilon}\right)$, independent of the number of agents $M$. The proof of Theorem \ref{thm:avg-policy-fed} is provided in Appendix~\ref{sec:analysis_policy}.

%% file: discussion.tex
\section{Conclusion}

We introduce a Q-learning algorithm for average-reward MDPs, applicable to both single-agent and federated settings. With carefully chosen learning rates and discount factors, our approach achieves notable improvements in sample complexity compared to previous Q-learning analyses for the single-agent case, and attains linear speedup in sample efficiency relative to the number of agents in the federated setting. Furthermore, our analysis establishes that the required number of communication rounds remains independent of the number of agents, thereby offering scalability.

%% file: analysis.tex
\section{Preliminaries}

\paragraph{Solutions to Bellman equation.} 
By \eqref{eq-bellman}, the optimal bias function $h^{\star}$ satisfies the Bellman optimality equation. For convenience, we restate it here:
\begin{align}\label{eq-bellman-supp}
J^{\star}+h^{\star}(s) = \max_{a\in\mathcal{A}}\left[r(s,a) + \sum_{s'\in\mathcal{S}}P(s'|s,a)h^{\star}(s')\right]=\max_{a\in\mathcal{A}}\left[r(s,a) + \mathbb{E}_{s'\sim P(\cdot|s,a)}[h^{\star}(s')]\right].
\end{align}
By introducing the bias Q-function $h_q^{\star}(s,a)$ defined as
\begin{align*}
h_q^{\star}(s,a) \coloneqq
r(s,a) + \mathbb{E}_{s'\sim P(\cdot|s,a)}[h^{\star}(s')] - J^{\star},
\end{align*}
we can rewrite \eqref{eq-bellman-supp} as
\begin{align}\label{eq:bellman-supp-2}
J^{\star}+h_q^{\star}(s,a) &= r(s,a) + \mathbb{E}_{s'\sim P(\cdot|s,a)}[h^{\star}(s')],\quad\mathrm{and}\\
\max_{a\in\mathcal{A}}h_q^{\star}(s,a)&= \max_{a\in\mathcal{A}}\left[r(s,a) + \mathbb{E}_{s'\sim P(\cdot|s,a)}[h^{\star}(s')]\right] - J^{\star} = h^{\star}(s).
\end{align}
Motivated by the observation that \eqref{eq:bellman-supp-2} is shift-invariant, we define normalized value functions $V^{\star}$ and $Q^{\star}$ by shifting $h^{\star}(s)$ and $h^{\star}_q(s,a)$ by a constant $c$: 
\begin{align}\label{eq:def-VQstar}
V^{\star}(s)\coloneqq h^{\star}(s) - c,\qquad Q^{\star}(s,a)\coloneqq h_q^{\star}(s,a) - c,\quad \mathrm{where}\quad c\coloneqq \frac{\max_{s'}h^{\star}(s')+\min_{s'}h^{\star}(s')}{2}.
\end{align}
This normalization ensures that $\|V^{\star}\|_\infty=\|h^{\star}\|_{\mathsf{span}}$.
One can then verify that  $(V^{\star},Q^{\star})$ satisfy the same Bellman relations, namely
\begin{align}\label{eq:bellman-vq}
J^{\star}+Q^{\star}(s,a) = r(s,a) + \mathbb{E}_{s'\sim P(\cdot|s,a)}[V^{\star}(s')],\quad\mathrm{and}\quad  \max_{a\in\mathcal{A}}Q^{\star}(s,a)=V^{\star}(s).
\end{align}

By virtue of \eqref{eq:bellman-vq}, one can straightforwardly verify that provided $\|h^{\star}\|_{\mathsf{span}}\gtrsim 1$, the $\ell_\infty$ norm of $V^{\star}$ and $Q^{\star}$ are controlled by the span semi-norm of the bias function:
\begin{align}\label{eq:proof-bound-VQstar}
\|V^{\star}\|_{\infty} = \|h^{\star}\|_{\mathsf{span}},\qquad \|Q^{\star}\|_{\infty}\le \|V^{\star}\|_{\infty} + \|r\|_{\infty} + J^{\star} \le \|h^{\star}\|_{\mathsf{span}} + 2\lesssim \|h^{\star}\|_{\mathsf{span}}.
\end{align}

\paragraph{Two auxiliary sequences.}
Next, we introduce two auxiliary sequences.
The first is based on the discounted value and Q functions $V_{\gamma_k}^{\star}(s)$ and $Q_{\gamma_k}^{\star}(s,a)$, $k=1,\ldots,K$, whose general definitions are given in \eqref{eq:def_V}.
Under the assumption that the AMDP is weakly communicating, Lemmas 6-8 of \citet{wang2022near} imply the following bounds:
\begin{lemma} \label{lemma:discount-avg-valuegap-span}
  If an AMDP is a weakly communicating MDP, for all $(s,a) \in \cS\times\cA$,
    \begin{align}
      | V_{\gamma}^{\star}(s) - J^{\star} | \le 4 (1-\gamma) \|h^{\star}\|_{\mathsf{sp}}
      ~~\text{and}~~
      | Q_{\gamma}^{\star}(s,a) - J^{\star} | \le 4 (1-\gamma) \|h^{\star}\|_{\mathsf{sp}}.
    \end{align}
\end{lemma}
Also, note that for any epoch $k$ and iteration $t$, and state $s \in \cS$,
\begin{align} \label{eq:value_gap_bound}
  |V_{k,t}(s) - V_{\gamma}^{\star}(s)| \le \max_{a \in \cA} |Q_{k,t}(s,a) - Q_{\gamma}^{\star}(s,a) | \le 1.
\end{align}

The second auxiliary sequence builds on the normalized functions $V^{\star}$ and $Q^{\star}$ from \eqref{eq:def-VQstar}. 
Define the sequence of value functions $\{V_k^{\star}\}_{k\ge 1}\subset\mathbb{R}^S$ by
\begin{align}\label{eq:def-Vkstar}
V_k^{\star}(s) \coloneqq {J}^{\star} + \frac1kV^{\star}(s),
\end{align}
which converges to $J^{\star}\bm{1}$ as $k\to\infty$.
Correspondingly, we introduce the Q-functions $Q_{k+1}^{\star}\in\mathbb{R}^{SA}$ via a Bellman-like update:
\begin{align}\label{eq:bellman-k}
Q_{k+1}^{\star}(s,a) \coloneqq \frac{1}{k+1}r(s,a) + \frac{k}{k+1} \mathbb{E}_{s'\sim P(\cdot|s,a)}[V_{k}^{\star}(s')],\qquad \forall (s,a)\in\mathcal{S}\times \mathcal{A},
\end{align}
with initialization $Q_1^{\star}(s,\pi^{\star}(s)) = V_1^{\star}(s)$ and $Q_1^{\star}(s,a) = 0$ for $a\neq\pi^{\star}(s)$.
By virtue of \eqref{eq:proof-bound-VQstar},
we obtain
\begin{align}\label{eq:bound-Vkstar}
\|V_k^{\star} - J^{\star}\|_{\infty} = \frac{1}{k}\|V^{\star}\|_{\infty}\le \frac{\|h^{\star}\|_{\mathsf{sp}}}{k}.
\end{align}
Furthermore, the sequence $Q_k^{\star}$ satisfies the following identities for all $k\ge 1$:
\begin{align}\label{eq:prop-Qstar}
Q_{k+1}^{\star}(s,a)&=J^{\star} + \frac{1}{k+1}Q^{\star}(s,a),\quad V_{k}^{\star}(s) = \max_{a\in\mathcal{A}}Q_{k}^{\star}(s,a),\quad \|Q_{k+1}^{\star}\|_{\infty}\le 1 + \frac{2+\|h^{\star}\|_{\mathsf{span}}}{k+1}.
\end{align}

\paragraph{Proof of \eqref{eq:prop-Qstar}.}
Substituting \eqref{eq:def-Vkstar} into \eqref{eq:bellman-k}, we obtain
\begin{align}\label{eq:proof-1}
Q_{k+1}^{\star}(s,a)&=\frac{1}{k+1}\left(r(s,a) + \mathbb{E}_{s'\sim P(\cdot|s,a)} [V^{\star}(s')]\right) + \frac{k}{k+1}J^{\star}\notag\\
&\overset{\eqref{eq:bellman-vq}}{=}\frac{1}{k+1}\left(J^{\star} + Q^{\star}(s,a)\right) + \frac{k}{k+1}J^{\star} = J^{\star} + \frac{1}{k+1}Q^{\star}(s,a),\qquad k\ge 1.
\end{align}
Moreover, we have
\begin{align*}
\max_{a\in\mathcal{A}}Q_{k}^{\star}(s,a) \overset{\eqref{eq:proof-1}}{=} \max_{a\in\mathcal{A}}\left[J^{\star} + \frac{1}{k}Q^{\star}(s,a)\right] = J^{\star} + \frac{1}{k}\max_{a\in\mathcal{A}}Q^{\star}(s,a) \overset{\eqref{eq:bellman-vq}}{=} J^{\star} + \frac{1}{k}V^{\star}(s)\overset{\eqref{eq:def-Vkstar}}{=}V_{k}^{\star}(s),\qquad \forall k\ge 1.
\end{align*}
Furthermore, by virtue of \eqref{eq:proof-bound-VQstar},
and \eqref{eq:proof-1}, we have
\begin{align*}
\|Q_{k+1}^{\star}\|_{\infty}\le \|J^{\star}\|_{\infty} + \frac{ \|Q^{\star}\|_{\infty}}{k+1} \le 1 + \frac{2+\|h^{\star}\|_{\mathsf{span}}}{k+1}.
\end{align*}

\paragraph{Useful properties of the learning rates.} The learning rates specified in \eqref{eq:para-group-1} and \eqref{eq:fed-para-group-1} have the following properties:
\begin{subequations} \label{eq:lr_prop}
\begin{align}
  \label{eq:lr_prop_1} \forall i \le t ~: &\quad \eta_{k,i} \prod_{j=i+1}^{t}(1-\eta_{k,j}) \le \eta_{k,t} \\
  \label{eq:lr_prop_2} \forall t\ge1 ~: &\quad \sum_{i=1}^t \eta_{k,i} \prod_{j=i+1}^{t}(1-\eta_{k,j}) + \prod_{j=1}^{t}(1-\eta_{k,j}) = 1 .
\end{align}
\end{subequations}
\textbf{Proof of \eqref{eq:lr_prop}.}
For the learning rates defined in \eqref{eq:para-group-1}, \eqref{eq:lr_prop_1} can be proved as follows:
\begin{align}
  \eta_{k,i} \prod_{j=i+1}^{t}(1-\eta_{k,j}) 
  &= \frac{1}{1+\frac{i^{2/3}}{8\log (4i)}} \frac{\frac{(i+1)^{2/3}}{8\log (4(i+1))}}{1+\frac{(i+1)^{2/3}}{8\log (4(i+1))}} \cdots \frac{\frac{t^{2/3}}{8\log (4t)}}{1+\frac{t^{2/3}}{8\log (4t)}} \cr
  &= \frac{\frac{(i+1)^{2/3}}{8\log (4(i+1))}}{1+\frac{i^{2/3}}{8\log (4i)}} \frac{\frac{(i+2)^{2/3}}{8\log (4(i+2))}}{1+\frac{(i+1)^{2/3}}{8\log (4(i+1))}} \cdots \frac{1}{1+\frac{t^{2/3}}{8\log (4t)}} \cr
  &\le \frac{1}{1+\frac{t^{2/3}}{8\log (4t)}} = \eta_{k,t},
\end{align}
where the last inequality follows from the fact that $(1+i)^{2/3} \le 1+i^{2/3}$ for any $i \ge 1$ due to subadditivity. Similarly, the property \eqref{eq:lr_prop_1} can be proved for the learning rates defined in \eqref{eq:fed-para-group-1}.
Next, we prove \eqref{eq:lr_prop_2} by induction. For $t=1$, it holds that
\begin{align}
  \sum_{i=1}^1 \eta_{k,i} \prod_{j=i+1}^{1}(1-\eta_{k,j}) + \prod_{j=1}^{1}(1-\eta_{k,j}) 
  = \eta_{k,1} + (1-\eta_{k,1}) = 1.
\end{align}
Then, suppose the statement holds for $t-1$, and we show it also holds for $t$:
\begin{align}
  \sum_{i=1}^t \eta_{k,i} \prod_{j=i+1}^{t}(1-\eta_{k,j}) + \prod_{j=1}^{t}(1-\eta_{k,j}) 
  &= \sum_{i=1}^{t-1} \eta_{k,i} \prod_{j=i+1}^{t-1}(1-\eta_{k,j})(1-\eta_{k,t}) + \eta_{k,t} + \prod_{j=1}^{t-1}(1-\eta_{k,j})(1-\eta_{k,t}) \cr
  &= (1-\eta_{k,t})\left(\sum_{i=1}^{t-1} \eta_{k,i} \prod_{j=i+1}^{t-1}(1-\eta_{k,j}) + \prod_{j=1}^{t-1}(1-\eta_{k,j})\right) + \eta_{k,t} \cr
  &= (1-\eta_{k,t})(1) + \eta_{k,t} = 1.
\end{align}

\paragraph{Notation.} In the remainder of the proof, we use vectors $V^{\star}, V_k^{\star}\in\mathbb{R}^S$ and $Q^{\star},Q_k^{\star}, r\in\mathbb{R}^{SA}$ to denote the respective mappings from $\mathcal{S}$ or $\mathcal{S}\times\mathcal{A}$ to $\mathbb{R}$. 
Let $P\in\mathbb{R}^{SA\times S}$ denote the transition probability matrix, where $P(s,a) = P(\cdot|s,a)$ is the probability vector corresponding to the state transition at the state-action pair $(s,a)$. In addition, we define the local empirical transition matrix at the $t$-th iteration in epoch $k$ in the single-agent setting as $P_{k,t} \in \{0,1\}^{SA \times S}$ as
  \begin{align} 
    P_{k,t}((s,a), s') \defn
    \begin{cases}
      1, \quad \text{if} ~ s' = s_{k,t}(s,a) \\
      0, \quad \text{otherwise}
    \end{cases}.
  \end{align}
In the federated setting, we denote the local empirical transition matrix for the $m$-th agent as $P_{k,t}^m \in \{0,1\}^{SA \times S}$ in a similar way.
Let $P^{\pi}\in\mathbb{R}^{S\times S}$ be the transition matrix under policy $\pi$, with the $s$-th row equal to $P(\cdot|s,\pi(s))$. Similarly, define $r^{\pi}\in\mathbb{R}^S$ as the reward vector under policy $\pi$, with the $s$-th entry given by $r(s,\pi(s))$.
For convenience, we write $P^{\star} = P^{\pi^{\star}}$ and $r^{\star} = r^{\pi^{\star}}$.
Notation $J^{\star}$ may refer to a scalar or to a vector with dimension implied by the context.
Finally, define the quantity $T_k\coloneqq\sum_{j=1}^k N_j$.
The inequalities $\le$ and $\ge$ between vectors are understood entry-wise.

Armed with these notation, update rules \eqref{eq:single-q-learning} in single-agent setting  and \eqref{eq:syncq-local} in federated setting are expressed by
\begin{subequations}
\begin{align}
Q_{k,t} &= (1-\eta_{k,t}) Q_{k,t-1} + \eta_{k,t} \bigg( (1-\gamma_k) r
 + \gamma_k P_{k,t} V_{\iota(k,t)} \bigg),\label{eq:single-q-learning-supp}\\
Q_{k,t}^m &= (1-\eta_{k,t}) Q_{k,t-1}^m + \eta_{k,t} \bigg( (1-\gamma_{k}) r
 + \gamma_{k} P_{k,t}^m  V_{k,\syn(k,t)}\bigg) ,\qquad  \forall 1\le m\le M. \label{eq:syncq-local-supp}
\end{align}
\end{subequations}

\section{Analysis in the single-agent setting (Theorem \ref{thm:average-reward})}
\label{sec:proof-single}
 
\subsection{Analysis for the first group of parameters}
\label{sec:proof-single-first}
For each epoch $k \in [K]$, we decompose the error as
\begin{align} \label{eq:error_decomp}
    \|Q_{k,t} - J^{\star}\|_{\infty} 
    \le \|Q_{\gamma_k}^{\star} - J^{\star}\|_{\infty} + \|Q_{k,t} - Q_{\gamma_k}^\star\|_{\infty}. 
\end{align}

The first error $\|Q_{\gamma_k}^{\star} - J^{\star}\|_{\infty}$, which arises from horizon mismatch, can be directly bounded according to Lemma~\ref{lemma:discount-avg-valuegap-span} as follows:
    \begin{align} \label{eq:err_horizon_mismatch}
    \|Q_{\gamma_k}^{\star} - J^{\star}\|_{\infty} 
    \le 3 (1-\gamma_k) \|h^{\star}\|_{\mathsf{sp}}.
\end{align}

Now, it suffices to show that the second error is bounded as
\begin{align}
  \|Q_{k,t} - Q_{\gamma_k}^\star\|_{\infty} \lesssim (1-\gamma_k) \|h^{\star}\|_{\mathsf{sp}}.
\end{align}

Denote $\Delta_{k,t} = Q_{k,t} - Q_{\gamma_k}^{\star}$. Since $\gamma_k$ remains fixed within stage $k$, the analysis follows a similar structure to the discounted MDP case in \citet{li2024q}. The key difference is that we derive error bounds in terms of both the discount factor and the span norm of the bias, ensuring that the second error converges at the same rate as the first error \eqref{eq:err_horizon_mismatch}, thereby balancing the convergence bottleneck.

\paragraph{Step 1: error decomposition.}
Using the fact that $Q_{\gamma_k}^{\star} = (1-\gamma_k)r + \gamma_k  P V_{\gamma_k}^{\star}$, we can write the error as
$$\Delta_{k,t} = (1-\eta_{k,t})\Delta_{k,t-1} + \eta_{k,t} \gamma_k \left( P_{k,t} V_{k,\syn(k,t)} - P V_{\gamma_k}^{\star}\right).$$
Then, it follows that
\begin{align} \label{eq:recursive_error_decomp_step1}
   \Delta_{k,t} 
  &= \prod_{i=1}^{t} (1-\eta_{k,i})\Delta_{k,0} + \gamma_k \sum_{i=1}^{t} \eta_{k,i}\left (\prod_{j=i+1}^{t}(1-\eta_{k,j}) \right )  \left( P_{k,i} V_{k,\syn(k,i)} - P V_{\gamma_k}^{\star}\right)  \cr
  &= \underbrace{\prod_{i=1}^{t} (1-\eta_{k,i})\Delta_{k,0} + \gamma_k \sum_{i=1}^{ t-g_k} \eta_{k,i} \prod_{j=i+1}^{t}(1-\eta_{k,j}) P_{k,i} \left(  V_{k,\syn(k,i)} - V_{\gamma_k}^{\star}\right)}_{=: E_{k,t}^1}\cr
    &\qquad + \underbrace{\gamma_k \sum_{i=1}^{t} \eta_{k,i} \prod_{j=i+1}^{t}(1-\eta_{k,j}) \left( P_{k,i}  - P \right) V_{\gamma_k}^{\star}}_{=: E_{k,t}^2}  \cr
    &\qquad + \underbrace{\gamma_k \sum_{i= t-g_k +1}^{t} \eta_{k,i} \prod_{j=i+1}^{t}(1-\eta_{k,j}) P_{k,i} \left(  V_{k,\syn(k,i)} - V_{\gamma_k}^{\star}\right)}_{=: E_{k,t}^3},
\end{align}
where $g_k \defn \left\lceil \frac{-\log(1-\gamma_k)^2}{\eta_{k,N_k}} \right\rceil$.

\paragraph{Step 2: bounding the decomposed errors.}
We can bound the error terms separately as follows:
\begin{itemize}
\item {Bounding the initialization error $E_{k,t}^1$.} For sufficiently large $t \ge g_k = \left\lceil \frac{-\log(1-\gamma_k)^2}{\eta_{k,N_k}} \right\rceil$, with the proposed learning rates \eqref{eq:lr_prop}, the initialization error decays at least in the rate of 
  \begin{align}
    \prod_{i=\lfloor t-g_k\rfloor+1}^{t}(1-\eta_{k,i}) \le (1-\eta_{k,t})^{g_k} \le\exp (-\eta_{k,t} g_k) \le (1-\gamma_k)^2.
  \end{align}
  Since $\|V_{k,\syn(k,i)} - V_{\gamma_k}^{\star}\|_{\infty} , ~\|\Delta_{k,0}\|_{\infty} \le 1$ according to \eqref{eq:value_gap_bound}, it follows that $\|E_{1}\|_{\infty} \le 2(1-\gamma_k)^2$.
\item {Bounding the transition variance $E_{k,t}^2$.}
    By applying Bernstein inequality, we obtain
    \begin{lemma} \label{lemma:e2}
    For any $( (k,t), s,a) \in [T_k] \times \cS\times\cA$ and $\delta\in (0,1)$, the following holds
    \begin{align}
        \|E_{k,t}^2\|_{\infty}
        &\lesssim (1-\gamma_k)^2 \|h^{\star}\|_{\mathsf{sp}} \log{\frac{2 |\cS||\cA|T_K}{\delta}}
    \end{align}
    at least with probability $1-\delta$, where $T_K = \sum_{k=1}^K N_k$.
    \end{lemma}
    \paragraph{Proof of Lemma~\ref{lemma:e2}.}
    Let $z_{k,i}$ be a random vector defined as $z_{k,i} = \eta_{k,i} \prod_{j=i+1}^{t}(1-\eta_{k,j}) \left( P_{k,i}  - P \right) V_{\gamma_k}^{\star} $, then
    \begin{align}
      E_{k,t}^2
        &= \sum_{i=1}^t  z_{k,i} .
    \end{align}    
    Since each $z_{k,i}$ is an independent random variable with zero mean, we aim to apply the Hoeffding inequality to bound $E_{k,t}^2$.
    Due to the properties of the learning rates \eqref{eq:lr_prop}, we first derive the following bounds:
    \begin{align}
        \sum_{i=1}^{t} \left(\eta_{k,i} \prod_{j=i+1}^{t}(1-\eta_{k,j}) \right)^2 \| V_{\gamma_k}^{\star} \|_{\mathsf{sp}}^2 
        \le \eta_{k,t}  \| V_{\gamma_k}^{\star} \|_{\mathsf{sp}}^2 
        \le 16 (1-\gamma_k)^2 (6(1-\gamma_k) \|h^{\star}\|_{\mathsf{sp}})^2 ,
    \end{align}
    where $\eta_{k,t} \le 16(1-\gamma_k)^2$ for $t \ge \frac{N_k}{2}$ and 
    $$\| V_{\gamma_k}^{\star} \|_{\mathsf{sp}} = \left|\max_{s}V_{\gamma_k}^{\star}(s) - \min_{s}V_{\gamma_k}^{\star}(s) \right| \le 6(1-\gamma_k) \|h^{\star}\|_{\mathsf{sp}}$$ according to Lemma~\ref{lemma:discount-avg-valuegap-span}. Applying the Hoeffding inequality yields
    \begin{align}
      \left \| \sum_{i=1}^t  z_{k,i}  \right \|_{\infty}
      &\le \sqrt{\sum_{i=1}^{t} \left(\eta_{k,i} \prod_{j=i+1}^{t}(1-\eta_{k,j}) \right)^2 \| V_{\gamma_k}^{\star} \|_{\mathsf{sp}}^2 \log{\frac{2|\cS||\cA|N}{\delta}}}\cr
        &\le  \sqrt{576 (1-\gamma_k)^4 \|h^{\star}\|_{\mathsf{sp}}^2 \log{\frac{2 |\cS||\cA|T_K}{\delta}} } \cr 
        &\le 24 (1-\gamma_k)^2 \|h^{\star}\|_{\mathsf{sp}} \log{\frac{2 |\cS||\cA|T_K}{\delta}},
    \end{align}
    where $T_K = \sum_{k=1}^K N_k$, and this completes the proof.
    
\item {Bounding the recursive optimality gap $E_{k,t}^3$.} 
  Using the fact that $\|V_{k,i} - V_{\gamma_k}^{\star}\|_{\infty} \le \|\Delta_{k,i}\|_{\infty}$, following from \eqref{eq:value_gap_bound},
  \begin{align}
    \|E_{k,t}^3\|_{\infty} \le \gamma_k \max_{\syn(k, t-g_k+1 )\le i <t} \|\Delta_{k,i}\|_{\infty}.
  \end{align}
\end{itemize}
Then, by combining the error bounds altogether, we obtain
\begin{align}
  \|\Delta_{k,t} \|_{\infty} \lesssim (1-\gamma_k)^2 \|h^{\star}\|_{\mathsf{sp}}  + \gamma_{k} \max_{t-2g_k\le i <t} \|\Delta_{k,i}\|_{\infty}
\end{align}
for sufficiently large $t \ge 2g_k = 2 \left\lceil \frac{-\log(1-\gamma_k)^2}{\eta_{k,N_k}}\right \rceil$.
Note that given $N_k \ge 1000$ for any $k\in[K]$, it follows $N_k \ge 2g_k$ since $(\eta_k)^{-1} \le 2 (N_k)^{2/3}$ and $(N_k)^{-1/3} \le (1-\gamma_k) \le 1$.

\paragraph{Step 3: solving recursion.} Now, we solve the recursive relation we obtained from the previous step as follows: 
\begin{align} \label{eq:err_discount_optimal_gap}
  \|\Delta_{k,N_k} \|_{\infty}
  &\lesssim (1-\gamma_k)^2 \|h^{\star}\|_{\mathsf{sp}} + \gamma_{k} \max_{N_k-2g_k\le i<N_k} \| \Delta_{k,i}\|_{\infty} \cr
  &\lesssim (1-\gamma_k)^2 \|h^{\star}\|_{\mathsf{sp}} + \gamma_{k} (1-\gamma_k)^2 \|h^{\star}\|_{\mathsf{sp}} + (\gamma_{k})^2 \max_{N_k-4g_k\le i<N_k} \| \Delta_{k,i}\|_{\infty} \cr
  &\lesssim \frac{1}{1-\gamma_k} (1-\gamma_k)^2 \|h^{\star}\|_{\mathsf{sp}} + (\gamma_{k})^L \max_{N_k-2Lg_k\le i<N_k} \| \Delta_{k,i}\|_{\infty}\cr
  &\lesssim  (1-\gamma_k) \|h^{\star}\|_{\mathsf{sp}} + e^{- (1-\gamma_{k})L} \cr
  &\lesssim (1-\gamma_k) \|h^{\star}\|_{\mathsf{sp}}
\end{align}
for $L = \left\lceil \frac{ -\log(1-\gamma_k)  }{(1-\gamma_k)} \right\rceil $ and $\frac{N_k}{2} \ge 2 \left(\frac{ \log(4/(1-\gamma_k))  }{(1-\gamma_k)}\right) \frac{ \log(1/(1-\gamma_k)^2)}{\eta_{k,N_k}} \ge 2 L g_k$ because the learning rates and discount factors in \eqref{eq:para-group-1} satisfy $(1-\gamma_k)^{-1} \le N_k $ and
\begin{align}
  \frac{ 2\log(4N_k)  }{(N_k)^{1/3}} \le (1-\gamma_k) ~~\text{and}~~
   \frac{ 4 \log(N_k)}{(N_k)^{2/3}}  \le \eta_{k,N_k}.
\end{align}
Finally, by plugging \eqref{eq:err_discount_optimal_gap} and \eqref{eq:err_horizon_mismatch} into \eqref{eq:error_decomp}, we conclude that after $K$ stages, the error of Q-estimates is bounded as
\begin{align}
    \|Q_{K,N_K} - J^{\star}\|_{\infty} 
    \lesssim (1-\gamma_k) \|h^{\star}\|_{\mathsf{sp}} \lesssim \frac{ \|h^{\star}\|_{\mathsf{sp}}}{(N_K)^{1/3}}. 
\end{align}

To achieve $\|Q_{K,N_K} - J^{\star}\|_{\infty}  \le \varepsilon$, $K$ should be large enough to satisfy
\begin{align}
  N_K = \max(1000, 2^K) \gtrsim \left(\frac{\|h^{\star}\|_{\mathsf{sp}}}{\varepsilon} \right)^3
\end{align}
and the total number of samples is bounded as
\begin{align}
  SA \sum_{k=1}^K N_k \le SA \left(10 + \sum_{k=1}^K 2^k \right) = \tilde{O}\left(SA \left(\frac{\|h^{\star}\|_{\mathsf{sp}}}{\varepsilon}\right)^3 \right).
\end{align}
Here, note that we need $N_k \ge 1000$ to make sure a large enough number of iterations is given, $N_k \ge 2g_k$ for small $k$, which is required for the error bounds in Step 2 to hold for all $k\in[K]$.

\subsection{Analysis for the second group of parameters}
\label{subsec:proof-thm-single-secondgroup}

\paragraph{Step 1: error decomposition.}
We begin from the $k$-th epoch, which aims to approximate $Q_{k+1}^{\star}$.
Combining the update rule \eqref{eq:single-q-learning-supp} and the definition of $Q^{\star}_{k+1}$ (cf. \eqref{eq:bellman-k}), we express the estimation error at the $t$-th step as:
\begin{align*}
Q_{k,t+1} - Q_{k+1}^{\star} 
&= (1-\eta_k)(Q_{k,t} - Q_{k+1}^{\star}) + \eta_k\left((1-\gamma_k)r + \gamma_kP_{k,t}V_{\iota(k,t)} - \frac{r}{k+1}-\frac{k}{k+1}PV_k^{\star}\right)\notag
\end{align*}
Note that $\iota(k,t) = (k-1,N_{k-1})$, and $\gamma_k = k/(k+1)$.
For ease of notation, we denote
$$
V_k \coloneqq V_{k,0} = V_{k-1,N_{k-1}}.
$$
The estimation error can be further simplified as:
\begin{align*}
Q_{k,t+1} - Q_{k+1}^{\star}&= (1-\eta_k)(Q_{k,t} - Q_{k+1}^{\star}) + \frac{k\eta_k}{k+1}(P_{k,t}V_{k} - PV_k^{\star})
\end{align*}
By recursion, the estimation error at the end of the $k$-th stage becomes:
\begin{align}\label{eq:proof-single-second-decomp-Q}
& Q_{k+1,0} - Q_{k+1}^{\star} = Q_{k,N_k} - Q_{k+1}^{\star}  \nonumber \\
&= (1-\eta_k)^{N_k}(Q_{k,0} - Q_{k+1}^{\star}) + \frac{k\eta_k}{k+1}\sum_{i=0}^{N_k-1}(1-\eta_k)^{N_k-1-i}(P_{k,i}V_{k} - PV_k^{\star})\notag\\
&=(1-\eta_k)^{N_k}(Q_{k,0} - Q_{k}^{\star}) + (1-\eta_k)^{N_k}(Q_{k}^{\star} - Q_{k+1}^{\star}) + \frac{k\eta_k}{k+1}\sum_{i=0}^{N_k-1}(1-\eta_k)^{N_k-1-i}(P_{k,i}V_{k} - PV_k^{\star}).
\end{align}

Define the error term as
$$
\Delta_{k+1}\coloneqq Q_{k+1,0} - Q_{k+1}^{\star} \in\mathbb{R}^{SA}.
$$
Identity \eqref{eq:proof-single-second-decomp-Q} then tells us that
\begin{align*}
\Delta_{k+1} = (1-\eta_k)^{N_k}\Delta_k + (1-\eta_k)^{N_k}(Q_{k}^{\star} - Q_{k+1}^{\star}) + \frac{k\eta_k}{k+1}\sum_{i=0}^{N_k-1}(1-\eta_k)^{N_k-1-i}(P_{k,i}V_{k} - PV_k^{\star})
\end{align*}
By recursion, $\Delta_{k+1}$ can be decomposed as:
\begin{align}\label{eq:Delta_k_decomp}
\Delta_{k+1}&=\underbrace{\prod_{j=1}^k(1-\eta_j)^{N_j}\Delta_1}_{=: E_{k}^1} + \underbrace{\sum_{j=1}^k \prod_{l=j}^k(1-\eta_l)^{N_l} (Q_j^{\star} - Q_{j+1}^{\star})}_{ =: E_{k}^2}\notag\\
&\quad + \sum_{j=1}^k \underbrace{\prod_{l=j+1}^k(1-\eta_l)^{N_l} \frac{j\eta_j}{j+1}\sum_{i=0}^{N_j-1}(1-\eta_j)^{N_j-1-i}(P_{j,i}V_{j} - PV_j^{\star})}_{=: E_{k,j}^3}.
\end{align}

\paragraph{Step 2: bounding the decomposed errors.}
Now we intend to bound the error terms in the right-hand-side of \eqref{eq:Delta_k_decomp} separately as follows: 

\begin{itemize}
\item {Bounding the initialization error $E_{k}^1$ and errors from historical epochs $E_k^2$ and $E_{k,j}^3$ for $j<k$.} 
Notice that as long as $N_k\ge 4\log T_k$, which is satisfied by $c_N\ge 4\log c_N + 3\log 2$, then the following decay factor at the $k$-th stage satisfies
\begin{align}\label{eq:proof-single-second-bound-eta}
(1-\eta_k)^{N_k}\le \exp(-N_k\eta_k) \overset{\text{(i)}}{=} \exp\left(-\frac{N_k}{1+\frac{N_k}{4\log T_k}}\right) \le \exp\left(-\frac{N_k}{\frac{N_k}{2\log T_k}}\right) = \frac{1}{T_k^2},
\end{align}
where (i) arises from the value of $\eta_{k,t}$ in \eqref{eq:para-group-2}.
Consequently, the first error term $E_k^1$ in the right-hand-side of \eqref{eq:Delta_k_decomp} is controlled by:
\begin{align}\label{eq:proof-single-second-bound-Ek1}
&\left\|E_k^1\right\|_{\infty} \le \frac{\|\Delta_1\|_{\infty}}{T_k^2} = \frac{\|Q^{\star}_1 - Q_{1,0}\|_{\infty}}{T_k^2} \overset{\text{(i)}}{=} \frac{\|Q^{\star}_1\|_{\infty}}{T_k^2} \overset{\text{(ii)}}{\le}  \frac{1+\|h^{\star}\|_{\mathsf{span}}}{T_k^2},
\end{align}
where (i) holds since $Q_{1,0}=Q_{0,0} = 0$, and (ii) uses property \eqref{eq:prop-Qstar}.
Similarly, using \eqref{eq:proof-single-second-bound-eta} yields the following bound for the error term $E_k^2$:
\begin{align}\label{eq:proof-single-second-bound-Ek2}
\|E_k^2\|_{\infty}&\le \sum_{j=1}^k\frac{\|Q_j^{\star}-Q_{j+1}^{\star}\|_{\infty}}{T_k^2} \overset{\text{(i)}}{\le}\frac1{T_k^2} \sum_{j=1}^k\left\|J^{\star}+\frac{Q^{\star}}{j}-J^{\star}-\frac{Q^{\star}}{j+1}\right\|_{\infty} \notag\\
&=  \frac{\|Q^{\star}\|_{\infty}}{T_k^2}\sum_{j=1}^k\frac{1}{j(j+1)} \overset{\text{(ii)}}{\le} \frac{\|Q^{\star}\|_{\infty}}{T_k^2}\overset{\text{(iii)}}{\le} \frac{2+\|h^{\star}\|_{\mathsf{span}}}{T_k^2},
\end{align}
where (i) holds since \eqref{eq:prop-Qstar}, (ii) arises from $\sum_{j=1}^k 1/(j(j+1)) = 1-1/(k+1)$, and (iii) uses the bound of $\|Q^{\star}\|_{\infty}$ given in \eqref{eq:proof-bound-VQstar}.

Moreover, the error terms $E_{k,j}^3$ for $j<k$ in the right-hand-side of \eqref{eq:Delta_k_decomp} satisfy:
\begin{align}
\left\|\sum_{j=1}^{k-1}E_{k,j}^3\right\|_{\infty}&=\left\|\sum_{j=1}^{k-1} \prod_{l=j+1}^k(1-\eta_l)^{N_l} \frac{j\eta_j}{j+1}\sum_{i=0}^{N_j-1}(1-\eta_j)^{N_j-1-i}(P_{j,i}V_{j} - PV_j^{\star})\right\|_{\infty} \notag\\
&\le \sum_{j=1}^{k-1} \prod_{l=j+1}^k(1-\eta_l)^{N_l} \eta_j\sum_{i=0}^{N_j-1}(1-\eta_j)^{N_j-1-i}(\|V_{j}\|_{\infty} + \|V_j^{\star}\|_{\infty})\notag\\
&\le \frac{1}{T_k^2}\sum_{j=1}^{k-1}(\|V_{j}\|_{\infty} + \|V_j^{\star}\|_{\infty}),
\end{align}
where the last inequality arises from \eqref{eq:proof-single-second-bound-eta} and $\eta_j\sum_{i=0}^{N_j-1}(1-\eta_j)^{N_j-1-i}\le \eta_j/\eta_j=1$.
By virtue of \eqref{eq:bound-Vkstar}, we have $\|V_j^{\star}\|_{\infty} \le 1+\|h^{\star}\|_{\mathsf{span}}/j$ and $\|V_j\|_{\infty}\le 1$,
which further yields
\begin{align}\label{eq:proof-single-second-bound-Ekj3}
\left\|\sum_{j=1}^{k-1}E_{k,j}^3\right\|_{\infty}&\le\frac{k(2+\|h^{\star}\|_{\mathsf{sp}})}{T_k^2} \lesssim \frac{k\|h^{\star}\|_{\mathsf{sp}}}{T_k^2},
\end{align}
provided that $\|h^{\star}\|_{\infty}\gtrsim 1$.

\item{Bounding the error term from the last epoch $E_{k,k}^3$.}
Note that the error term $E_{k,k}^3$ does not contain the factor $(1-\eta_k)^{N_k}$, and thus its $\ell_{\infty}$ norm cannot be bounded in the same manner as \eqref{eq:proof-single-second-bound-Ekj3}
To control $\|E_{k,k}^3\|_{\infty}$, we need to establish a new recursive expression.
Inserting \eqref{eq:proof-single-second-bound-Ek1}, \eqref{eq:proof-single-second-bound-Ek2}, and \eqref{eq:proof-single-second-bound-Ekj3} into \eqref{eq:Delta_k_decomp}, we bound $\Delta_{k+1}$ as: 
\begin{align}
\Delta_{k+1} 
& = \frac{k\eta_k}{k+1}\sum_{i=0}^{N_k-1}(1-\eta_k)^{N_k-1-i}(P_{k,i}V_{k} - PV_k^{\star}) + O\left(\frac{k\|h^{\star}\|_{\mathsf{sp}}}{T_k^2}\right)\notag\\
&=\frac{k\eta_k}{k+1}\sum_{i=0}^{N_k-1}(1-\eta_k)^{N_k-1-i}P(V_{k} - V_k^{\star}) + \frac{k\eta_k}{k+1}\sum_{i=0}^{N_k-1}(1-\eta_k)^{N_k-1-i}(P_{k,i}-P)V_{k} + O\left(\frac{k\|h^{\star}\|_{\mathsf{sp}}}{T_k^2}\right),\label{eq:Delta_k_decomp_2}
\end{align}
where the first term reflects the bias and the second term comes from the randomness of sampling.

Let $\pi_k(s) \coloneqq \arg\max_{a\in\mathcal{A}} Q_k(s,a)$.
Recalling that $P(V_{k} - V_k^{\star})=P^{\pi_k}Q_{k} - P^{\pi_{\star}}Q_k^{\star} \le P^{\pi_k}(Q_{k} - Q_k^{\star}) = P^{\pi_k}\Delta_{k}$, where the $\le$ is entry-wise, we apply recursion to \eqref{eq:Delta_k_decomp_2} and obtain
\begin{align}\label{eq:Delta_k_decomp_3}
\Delta_{k+1} 
&\le \frac{k\alpha_k}{k+1} P^{\pi_{k}}\Delta_k  + \frac{k\eta_k}{k+1}\sum_{i=0}^{N_k-1}(1-\eta_k)^{N_k-1-i}(P_{k,i}-P)V_{k} + 
O\left(\frac{k\|h^{\star}\|_{\mathsf{sp}}}{T_k}\right)\notag\\
&\le \underbrace{\frac{1}{k+1} \prod_{l=1}^k(\alpha_lP^{\pi_l})\Delta_1}_{=:E_{k,k,0}^{3}} \notag\\
& \quad + \sum_{j=1}^k\prod_{l=j+1}^k(\alpha_lP^{\pi_l})\frac{j}{k+1}\underbrace{\eta_j\sum_{i=0}^{N_j-1}(1-\eta_j)^{N_j-1-i}(P_{j,i}-P)V_{j}}_{=: E_{k,k,j}^{3}} + 
O\left(\frac{\|h^{\star}\|_{\mathsf{sp}}}{(k+1)c_N^2}\right),
\end{align}
where $\alpha_k \coloneqq \eta_k\sum_{i=0}^{N_k-1}(1-\eta_k)^{N_k-1-i}\le 1$, and the last inequality uses $T_k\ge N_k \ge c_N k^2$. Noting that $\|\Delta_1\|_{\infty} \le 1+ \|h^{\star}\|_{\mathsf{sp}}$ (cf. \eqref{eq:proof-single-second-bound-Ek1}) and the $\ell_1$ norm of probability transition matrix $\|P^{\pi_l}\|_1= 1$,
we have
\begin{align}\label{eq:proof-single-second-bound-Ekk03}
\|E_{k,k,0}^3\|_{\infty}\le \frac{1+\|h^{\star}\|_{\mathsf{sp}}}{k+1}.
\end{align}
Next, we shall control $E_{k,k,j}^3$. 
Note that
\begin{align}\label{eq:proof-single-second-sp-Vk-Vbar}
\|V_{k} - J^{\star}\|_{\mathsf{sp}}
\le\|V_{k} - V_k^{\star}\|_{\mathsf{sp}} + \|V_k^{\star}-J^{\star}\|_{\mathsf{sp}}
\overset{\text{(i)}}{=} \|V_{k} - V_k^{\star}\|_{\mathsf{sp}} + \frac{\|V^{\star}\|_{\mathsf{sp}}}{k} \le \left\|\Delta_k\right\|_{\infty} + \frac{\|h^{\star}\|_{\mathsf{sp}}}{k},
\end{align}
where (i) comes from the fact that
$
V_{k}^{\star} - J^{\star} = \frac{V^{\star}}{k}
$ (cf. \eqref{eq:def-Vkstar}).
Moreover, notice that all entries in $J^{\star}$ are identical, the variance of the second term $E_{k,k,j}^3$ is bounded by
\begin{align}\label{eq:proof-single-second-varEkkj3}
\mathsf{Var}\left(E_{k,k,j}^3\right) 
&= \mathsf{Var}\left(\eta_j\sum_{i=0}^{N_j-1}(1-\eta_j)^{N_j-1-i}(P_{j,i}-P)(V_{j}-J^{\star})\right)\notag\\
&\le \frac{\eta_j^2}{1-(1-\eta_j)^2} \mathsf{Var}\left((P_{j,i}-P)(V_{j}-J^{\star})\right)\notag\\
&\le \eta_j \|V_j - J^{\star}\|_{\mathsf{sp}}^2\notag\\
&\overset{\text{(i)}}{\le} \eta_j \left(\|\Delta_j\|_{\infty} + \frac{\|h^{\star}\|_{\mathsf{sp}}}{j}\right)^2,
\end{align}
where (i) comes from \eqref{eq:proof-single-second-sp-Vk-Vbar}.
By applying Bernstein's inequality, with probability at least $1-\delta/(2j^2)$, the following holds; moreover, by a union bound over all $j\in[K]$, the overall probability is at least $1-\delta$:
\begin{align}
\|E_{k,k,j}^3\|_{\infty}
&\lesssim \sqrt{\mathsf{Var}(E_{k,k,j}^3)\log\frac{2SAj^2}{\delta}} + \eta_j\|V_j-J^{\star}\|_{\infty}\log\frac{2SAj^2}{\delta}\notag\\ 
&\overset{\text{}}{\lesssim}  \sqrt{\eta_j}\left(\|\Delta_j\|_{\infty} + \frac{\|h^{\star}\|_{\mathsf{sp}}}{j}\right)\sqrt{\log\frac{SAj}{\delta}}\label{eq:proof-2}\\
&\overset{\text{(i)}}{\lesssim} \sqrt{\frac{\log T_j}{N_j}}\left(\|\Delta_j\|_{\infty} + \frac{\|h^{\star}\|_{\mathsf{sp}}}{j}\right)\sqrt{\log\frac{SAj}{\delta}}\notag\\
&\overset{\text{(ii)}}{\lesssim}  \frac{1}{j\log (j+1)}\left(\|\Delta_j\|_{\infty} + \frac{\|h^{\star}\|_{\mathsf{sp}}}{j}\right)\frac{1}{\sqrt{c_N}},\label{eq:proof-single-second-bound-Ekkj3}
\end{align}
where (i) comes from the definition of $\eta_j$ (cf. \eqref{eq:para-group-2}):
\begin{align}\label{eq:proof-single-second-bound-Ekkj3-explain1}
\eta_j = \frac{1}{1+\frac{N_j}{4\log (\sum_{i=1}^jN_i)}} = \frac{1}{1+\frac{N_j}{4\log T_j}}\le \frac{4\log T_j}{N_j},
\end{align}
(ii) arises from 
\begin{align}\label{eq:proof-single-second-bound-Ekkj3-explain2}
N_j&= c_Nj^2\log^5 (j+1)\log^3\left(\frac{SA}{\delta}\right)\ge c_Nj^2\log^2(j+1)\left(\log (j+1)\log\left(\frac{SA}{\delta}\right)\right)^2\notag\\
&{\gtrsim} c_Nj^2\log^2(j+1)\log T_j\log\frac{SAj}{\delta},
\end{align}
by virtue of the fact that $\log T_j\lesssim \log (j^3\log^3(j+1)\log^3(SA/\delta))\lesssim \log(j+1) + \log(SA/\delta)\lesssim\log(j+1)\log(SA/\delta)$.
Inequality \eqref{eq:proof-2} holds since
\begin{align*}
\eta_k \le \frac{4\log T_j}{N_j}\lesssim \frac{\log T_j}{c_N j^2\log^2(j+1)\log T_j \log\frac{SAj}{\delta}} \lesssim \frac{1}{\log\frac{SAj}{\delta}}.
\end{align*}
Substituting \eqref{eq:proof-single-second-bound-Ekk03} and \eqref{eq:proof-single-second-bound-Ekkj3} into \eqref{eq:Delta_k_decomp_3}, provided that $\|h^{\star}\|_{\mathsf{span}}\gtrsim 1$, we have
\begin{align}\label{eq:proof-single-second-Delta-upper}
\Delta_{k+1}
&\lesssim\frac{\|h^{\star}\|_{\mathsf{sp}}}{k+1} + \frac{1}{k+1}\sum_{j=1}^k\frac{\theta}{\log (j+1)}\left(\|\Delta_j\|_{\infty}+\frac{\|h^{\star}\|_{\mathsf{sp}}}{j}\right) + \frac{\|h^{\star}\|_{\mathsf{sp}}}{(k+1)c_N^2}\notag\\
&\overset{\text{(i)}}{\lesssim} \frac{\|h^{\star}\|_{\mathsf{sp}}}{k+1} + \frac{1}{k+1}\sum_{j=1}^k\frac{\theta\|\Delta_j\|_{\infty}}{\log(j+1)} + \frac{\theta \|h^{\star}\|_{\mathsf{sp}}\log\log(k+1)}{k+1} \notag \\
& \asymp \frac{\|h^{\star}\|_{\mathsf{sp}}\log\log(k+1)}{k+1} + \frac{1}{k+1}\sum_{j=1}^k\frac{\theta\|\Delta_j\|_{\infty}}{\log(j+1)},
\end{align}
where $\theta \coloneqq 1/\sqrt{c_N}\ll 1$, and (i) uses the fact that $\sum_{j=1}^kj^{-1}\log^{-1}(j+1)\le 2\log\log (k+1)$.
\end{itemize}

Similarly, we can derive the following lower bound for $\Delta_{k+1}$:
\begin{align*}
\Delta_{k+1} 
&\gtrsim -\frac{\|h^{\star}\|_{\mathsf{sp}}\log\log(k+1)}{k+1} - \frac{1}{k+1}\sum_{j=1}^k\frac{\theta\|\Delta_j\|_{\infty}}{\log(j+1)}.
\end{align*}

\paragraph{Step 3: solving recursion.} 
Combining the upper and the lower bound, we control the infinite norm of $\Delta_{k+1}$ as
\begin{align}\label{eq:proof-singleagent-recursion}
\frac{\|\Delta_{k+1}\|_{\infty}}{\log(k+2)}
&\le \frac{c \|h^{\star}\|_{\mathsf{sp}} \log \log (k+1)}{(k+1)\log(k+2)} + \frac{c\theta }{(k+1)\log(k+2)}\sum_{j=1}^k\frac{\|\Delta_j\|_{\infty}}{\log(j+1)}\notag\\
&\le \frac{c \|h^{\star}\|_{\mathsf{sp}} }{k+1} + \frac{c\theta \Delta_k^{\mathsf{sum}}}{(k+1)\log(k+2)},
\end{align}
where $\Delta_k^{\mathsf{sum}} \coloneqq \sum_{j=1}^k\frac{\|\Delta_j\|_{\infty}}{\log(j+1)}$.

Now, we solve the recursive relation \eqref{eq:proof-singleagent-recursion}.
We first claim that for sequence $a_k$ satisfying 
$$
a_{k+1} = \lambda_k s_k +  \beta_k,
$$
where $\lambda_k,\beta_k,a_1 \ge 0$ and $s_k = \sum_{j=1}^k a_j$,  we have 
$$
a_{k+1}= \beta_{k} + \lambda_{k}\sum_{i=1}^{k-1}\prod_{j=i+1}^{k-1}(1+\lambda_j)\beta_i + \lambda_k\prod_{i=1}^{k-1}(1+\lambda_i)a_1.
$$
Taking $a_k \coloneqq \|\Delta_k\|/\log (k+1)$, $\lambda_k \coloneqq c\theta/(k+1)/\log(k+2)$, and $\beta_k \coloneqq c\|h^{\star}\|_{\mathsf{sp}}/(k+1)$,
we have
\begin{align}\label{eq:bound-Delta}
\frac{\|\Delta_{k+1}\|_{\infty}}{\log(k+2)}
&\le \frac{c\|h^{\star}\|_{\mathsf{sp}}}{k+1} + \frac{c\theta}{(k+1)\log(k+2)}\sum_{i=1}^{k-1}\prod_{j=i+1}^{k-1}\left(1+\frac{c\theta}{(j+1)\log(j+2)}\right)\frac{c\|h^{\star}\|_{\mathsf{sp}}}{i+1}\notag\\
&\quad + \frac{c\theta}{(k+1)\log(k+2)}\prod_{i=1}^{k-1}\left(1+\frac{c\theta}{(i+1)\log(i+2)}\right)\frac{\|\Delta_1\|_{\infty}}{\log 2}.
\end{align}
For $\theta\le 1/2$, the product in the right-hand-side of \eqref{eq:bound-Delta} satisfies 
\begin{align*}
\prod_{j=i+1}^{k-1}\left(1+\frac{\theta}{(j+1)\log(j+2)}\right)
&\le \exp\left(\sum_{j=i+1}^{k-1}\frac{\theta}{(j+1)\log(j+2)}\right) \notag\\
&\le \exp(2\theta\log\log(k+1)) \le \log k.
\end{align*}
Substituting it into \eqref{eq:bound-Delta}, we have
\begin{align}\label{eq:bound-Delta-2}
\frac{\|\Delta_{k+1}\|_{\infty}}{\log(k+2)}
&\le \frac{c\|h^{\star}\|_{\mathsf{sp}}}{k+1} + \frac{c^2\theta \|h^{\star}\|_{\mathsf{sp}}\log k}{(k+1)\log(k+2)}\sum_{i=1}^{k-1}\frac{1}{i+1} + \frac{c\theta\log k}{(k+1)\log(k+2)}\frac{\|\Delta_1\|_{\infty}}{\log 2}\notag\\
&\lesssim \frac{(1+\theta\log k)\|h^{\star}\|_{\mathsf{sp}}}{k+1}\lesssim \frac{\|h^{\star}\|_{\mathsf{sp}}\log k}{k+1}.
\end{align}
Recalling the definition of $\Delta_K$, and by virtue of $\|J^{\star} - Q_K^{\star}\|_{\infty} = \|Q^{\star}\|_{\infty}/K \lesssim \|h^{\star}\|_{\mathsf{sp}}/(K+1)$ (cf. \eqref{eq:prop-Qstar} and \eqref{eq:proof-bound-VQstar}),  we get the final result that
\begin{align}\label{eq:diff-JK-J}
\|Q_{K,N_K} - J^{\star}\|_{\infty} \le \|Q_{K,N_K} - Q_K^{\star}\|_{\infty} + \|J^{\star} - Q_K^{\star}\|_{\infty} \le \left\|\Delta_{K}\right\|_{\infty} + \frac{\|h^{\star}\|_{\mathsf{sp}}}{K}\lesssim \frac{\|h^{\star}\|_{\mathsf{sp}}\log K}{K}.
\end{align}

%% file: analysis_fl.tex
\section{Analysis for the federated setting (Theorem \ref{thm:average-reward-fed})}
\label{sec:proof-federated}

\subsection{Analysis for the first group of parameters}

The proof will follow similar steps as in the proof of the single-agent case in Section~\ref{sec:proof-single-first}. We omit some of the repetitive derivations and only highlight the key differences here.
Similar to the single-agent case, we split the error into two components, one due to horizon mismatch and the other due to the optimality gap in the discounted MDP, as follows:
\begin{align} \label{eq:error_decomp_fed}
    \|Q_{k,t} - J^{\star}\|_{\infty} 
    \le \|Q_{\gamma_k}^{\star} - J^{\star}\|_{\infty} + \|Q_{k,t} - Q_{\gamma_k}^\star\|_{\infty}. 
\end{align}
Since the first term can be bounded by $\widetilde{O}((1-\gamma_k) \|h^{\star}\|_{\mathsf{sp}})$ as \eqref{eq:err_horizon_mismatch} for the single-agent case, we only need to focus on proving the bound of the second error term, that is,
$
    \|Q_{k,t} - Q_{\gamma_k}^\star\|_{\infty} \lesssim (1-\gamma_k) \|h^{\star}\|_{\mathsf{sp}} 
$ for the first parameter group defined under the federated setup.
Since $\gamma_k$ remains fixed within stage $k$, the analysis follows a similar structure to the discounted MDP case in \citet{woo2023blessing}. The key difference is that we derive error bounds in terms of both the discount factor and the span norm of the bias, ensuring that the two decomposed error temrms in \eqref{eq:error_decomp} converges at the same rate, thereby balancing the convergence bottleneck.

\paragraph{Step 1: error decomposition.}
Unlike the single-agent case, where the error is defined in terms of a single Q-estimate, in the federated setting, the error at iteration $t$ of stage $k$ is defined in terms of averaged Q-estimates of all agents as follows: 
$$\Delta_{k,t} = \sum_{m=1}^M Q_{k,t}^m - Q_{\gamma_k}^{\star}.$$
Recalling the error decomposition for the single Q-estimate in \eqref{eq:recursive_error_decomp_step1} shown for the single-agent case, we decompose $\Delta_{k,t}$ as follows:
\begin{align}
   \Delta_{k,t}
  &= \prod_{i=1}^{t} (1-\eta_{k,i})\Delta_{k,0} + \frac{\gamma_k}{M} \sum_{i=1}^{t} \eta_{k,i} \prod_{j=i+1}^{t}(1-\eta_{k,j}) \sum_{m=1}^M \left( P_{k,i}^m( V_{k,\syn(k,i)} - P V_{\gamma_k}^{\star}\right)  \cr
  &= \underbrace{\prod_{i=1}^{t} (1-\eta_{k,i})\Delta_{k,0} + \frac{\gamma_k}{M} \sum_{i=1}^{ t-g_k } \eta_{k,i} \prod_{j=i+1}^{t}(1-\eta_{k,j}) \sum_{m=1}^M P_{k,i}^m \left(  V_{k,\syn(k,i)} - V_{\gamma_k}^{\star}\right)}_{ =: E_{k,t}^1}\cr
    &\qquad + \underbrace{\frac{\gamma_k}{M} \sum_{i=1}^{t} \eta_{k,i} \prod_{j=i+1}^{t}(1-\eta_{k,j}) \sum_{m=1}^M \left( P_{k,i}^m  - P \right) V_{\gamma_k}^{\star}}_{ =: E_{k,t}^2}  \cr
    &\qquad + \underbrace{\frac{\gamma_k}{M} \sum_{i= t-g_k  +1}^{t} \eta_{k,i} \prod_{j=i+1}^{t}(1-\eta_{k,j}) \sum_{m=1}^M P_{k,i}^m \left(  V_{k,\syn(k,i)} - V_{\gamma_k}^{\star}\right)}_{ =: E_{k,t}^3}.
\end{align}

\paragraph{Step 2: bounding the decomposed errors.}
With the proposed learning rates \eqref{eq:fed-para-group-1}, for any $k \in [K]$ and sufficiently large $2g_k \le t \le N_k $, we can derive the bound of the error terms as follows:
\begin{align}
    \|E_{k,t}^1\|_{\infty} &\le 2(1-\gamma_k)^2,  \\
    \|E_{k,t}^2\|_{\infty} &\le 24 (1-\gamma_k)^2 \|h^{\star}\|_{\mathsf{sp}} \log{\frac{2 |\cS||\cA|T_K}{\delta}},  \\  \label{eq:bound_e2_fed}
  \|E_{k,t}^3\|_{\infty} &\le \gamma_k \max_{\syn(k, t-g_k+1)\le i <t} \|\Delta_{k,i}\|_{\infty}.
\end{align}
Note that given $N_k \ge 1000$ for any $k\in[K]$, $N_k \ge 2g_k$ since $(\eta_k)^{-1} \le 2 (N_k)^{2/3} M^{-1/3}$ and $(MN_k)^{-1/3} \le (1-\gamma_k) \le 1$.
We omit the detailed derivations for the bounds of $E_{k,t}^1$ and $E_{k,t}^3$ since they follow similarly as in the single-agent case. See Section~\ref{sec:proof-single-first} step 2 for reference. The bound of the transition variance term $E_{k,t}^2$ is derived by similarly applying the Hoeffding bound, but the intermediate values are different from the single-agent case and are expressed in terms of the number of agents $M$, which appears only in the federated setting. We provide the detailed proof below.

    \paragraph{Proof of \eqref{eq:bound_e2_fed}.}
    Rewrite the second error term as the sum of random vector 
    $
      E_{k,t}^2 = \sum_{i=1}^t  z_{k,i},
    $
  where $z_{k,i} = \eta_{k,i} \prod_{j=i+1}^{t}(1-\eta_{k,j}) \left( P_{k,i}  - P\right) V_{\gamma_k}^{\star} $ is an independent random vector with zero mean.
    Due to the properties of the learning rates \eqref{eq:lr_prop}, we can derive the following bounds for $z_{k,i}$,
    \begin{align}
        \frac{1}{M^2} \sum_{m=1}^M \sum_{i=1}^{t} \left(\eta_{k,i} \prod_{j=i+1}^{t}(1-\eta_{k,j}) \right)^2 \| V_{\gamma_k}^{\star} \|_{\mathsf{sp}}^2 
        \le \frac{1}{M} \eta_{k,t}  \| V_{\gamma_k}^{\star} \|_{\mathsf{sp}}^2 
        \le 16 (1-\gamma_k)^2 (6(1-\gamma_k) \|h^{\star}\|_{\mathsf{sp}})^2 ,
    \end{align}
    where $\eta_{k,t} \le 16 M (1-\gamma_k)^2$ for $t \ge \frac{N_k}{2}$ and $\| V_{\gamma_k}^{\star} \|_{\mathsf{sp}} = |\max_{s}V_{\gamma_k}^{\star}(s) - \min_{s}V_{\gamma_k}^{\star}(s)| \le 6(1-\gamma_k) \|h^{\star}\|_{\mathsf{sp}}$ according to Lemma~\ref{lemma:discount-avg-valuegap-span} and the following properties of the learning rates
    \note{
      \begin{align}
        \eta_{k,t} = \frac{1}{1+\frac{(Mt)^{2/3}}{8M\log (4Mt)}} \le \frac{8M\log (4Mt)}{(Mt)^{2/3}} \le 16 M \frac{\log (2 M N_k)}{(M N_k)^{2/3}} \le 16 M (1-\gamma_k)^2. 
        \end{align}
    }    
    Since each $z_{k,i}$ is an independent random vector with zero mean, we apply the Hoeffding inequality as follows:
    \begin{align}
      \left| E_{k,t}^2 \right|
      &\le \sqrt{ \frac{1}{M^2} \sum_{m=1}^M \sum_{i=1}^{t} \left(\eta_{k,i} \prod_{j=i+1}^{t}(1-\eta_{k,j}) \right)^2 \| V_{\gamma_k}^{\star} \|_{\mathsf{sp}}^2 \log{\frac{2|\cS||\cA|T_K}{\delta}}}\cr
        &\le  \sqrt{576 (1-\gamma_k)^4 \|h^{\star}\|_{\mathsf{sp}}^2 \log{\frac{2 |\cS||\cA|T_K}{\delta}} } \cr 
        &\le 24 (1-\gamma_k)^2 \|h^{\star}\|_{\mathsf{sp}} \log{\frac{2 |\cS||\cA|T_K}{\delta}}
    \end{align}
    where $T_K = \sum_{k=1}^K N_k$, and this completes the proof.

\paragraph{Step 3: solving recursion.} Combining the error bounds obtained  from the previous steps, we solve the recursive relation as follows: 
\begin{align} \label{eq:final_error_fed}
  \|\Delta_{k,N_k} \|_{\infty}
  &\lesssim \frac{1}{1-\gamma_k} (1-\gamma_k)^2 \|h^{\star}\|_{\mathsf{sp}} + (\gamma_{k})^L \max_{N_k-2Lg_k\le i<N_k} \| \Delta_{k,i}\|_{\infty}\cr
  &\lesssim  (1-\gamma_k) \|h^{\star}\|_{\mathsf{sp}} + e^{- (1-\gamma_{k})L} \cr
  &\lesssim (1-\gamma_k) \|h^{\star}\|_{\mathsf{sp}}
\end{align}
for $L = \lceil \frac{ -\log(1-\gamma_k)  }{(1-\gamma_k)} \rceil $ and \note{$\frac{N_k}{2} \ge 2 (\frac{ \log(4/(1-\gamma_k))  }{(1-\gamma_k)}) \frac{ \log(1/(1-\gamma_k)^2)}{\eta_{k,N_k}} \ge 2 L g_k$} because the learning rates and discount \note{factors in \eqref{eq:fed-para-group-1} satisfy $(1-\gamma_k)^{-1} \le M N_k $ and
\begin{align}
  \frac{ 2\log(4M N_k)  }{(M N_k)^{1/3}} \le (1-\gamma_k) ~~\text{and}~~
   \frac{ 4 M \log(M N_k)}{(M N_k)^{2/3}}  \le \eta_{k,N_k}.
\end{align}}
Finally, by plugging \eqref{eq:final_error_fed} into \eqref{eq:error_decomp_fed}, we conclude that after $K$ stages, the error of averaged Q-estimate is bounded as
\begin{align}
    \|Q_{K,N_K} - J^{\star}\|_{\infty} 
   \lesssim  (1-\gamma_k) \|h^{\star}\|_{\mathsf{sp}} 
   \lesssim \frac{ \|h^{\star}\|_{\mathsf{sp}}}{(M N_K)^{1/3}}. 
\end{align}

To achieve $\|Q_{K,N_K} - J^{\star}\|_{\infty}  \le \varepsilon$, $K$ should be large enough to satisfy
\begin{align}
  N_K = \max(1000, 2^K) \gtrsim \frac{1}{M}\left(\frac{\|h^{\star}\|_{\mathsf{sp}}}{\varepsilon}\right)^3
\end{align}
and the total number of samples is bounded as
\begin{align}
  SA \sum_{k=1}^K N_k  = \tilde{O}\left(\frac{SA}{M} \left(\frac{\|h^{\star}\|_{\mathsf{sp}}}{\varepsilon}\right)^3\right).
\end{align}

\paragraph{Analysis of the number of communication rounds.}
When communication intervals at stage $k$ are set as $g_k$, \note{i.e., $ \min_{t_1 \neq t_2 \in \synset(k)} |t_1-t_2| = g_k$}, the number of communication rounds is bounded as
\begin{align}
  \synset(k) \lesssim \frac{N_k}{g_k}
               \le N_k \eta_{k,N_k} \le (M N_k)^{\frac{1}{3}}
\end{align}
Thus, the total number of communication rounds for $K$ stages is bounded by
\begin{align}
  \sum_{k=1}^K |\synset(k)|
  \lesssim M^{1/3} \sum_{k=1}^K (N_k)^{1/3} 
    \lesssim M^{1/3} K^{2/3}(\sum_{k=1}^K N_k)^{1/3} 
    \lesssim (MT_K)^{1/3},
\end{align}
since $2^K \le T_K = \sum_{k=1}^K N_k \le  10^4 + 2^{K+1} $ and $K = O(\log(T_k))$.

\subsection{Analysis for the second group of parameters}

The proof is similar to the argument in Section \ref{subsec:proof-thm-single-secondgroup}.
We omit some of the repetitive derivations and only highlight the key differences here.

\paragraph{Step 1: error decomposition.}
Similar to the derivation of \eqref{eq:proof-single-second-decomp-Q}, we split the error $\Delta_{k+1} \coloneqq Q_{k+1,0}^m-Q_{k+1}^{\star}$ as follows:
\begin{align}
Q_{k+1,0} - Q_{k+1}^{\star} 
&=(1-\eta_k)^{N_k}(Q_{k,0} - Q_{k}^{\star}) + (1-\eta_k)^{N_k}(Q_{k}^{\star} - Q_{k+1}^{\star})\notag\\
&\quad + \frac{1}{M}\frac{k\eta_k}{k+1}\sum_{m=1}^M\sum_{i=0}^{N_k-1}(1-\eta_k)^{N_k-1-i}(P_{k,i}^mV_{k} - PV_k^{\star}).
\end{align}
Using a similar recursive approach as in \eqref{eq:Delta_k_decomp}, we obtain the following counterpart:
\begin{align}
\Delta_{k+1}&=\underbrace{\prod_{j=1}^k(1-\eta_j)^{N_j}\Delta_1}_{=: E_{k}^1} + \underbrace{\sum_{j=1}^k \prod_{l=j}^k(1-\eta_l)^{N_l} (Q_j^{\star} - Q_{j+1}^{\star})}_{=:E_{k}^2}\notag\\
&\quad + \sum_{j=1}^k \underbrace{\prod_{l=j+1}^k(1-\eta_l)^{N_l} \frac{j\eta_j}{(j+1)M}\sum_{m=1}^{M}\sum_{i=0}^{N_j-1}(1-\eta_j)^{N_j-1-i}(P_{j,i}^mV_{j} - PV_j^{\star})}_{=:E_{k,j}^3}.
\end{align}
We claim that for sufficiently large $c_N$, the following inequalities hold, whose proof is postponed to the end of this section:
\begin{align}
N_k&\ge \log(MT_k),\qquad k\ge 1,\label{eq:proof-fl-second-claim-1}\\
MN_k& \gtrsim c_N k^2\log^2(k+1)\log\frac{SA}{\delta}\log(MT_k), \qquad k\ge 1.\label{eq:proof-fl-second-claim}
\end{align}
Based on this claim, we have
$
(1-\eta_k)^{N_k}\le \exp\left(-\frac{N_k}{\frac{N_k}{2\log (MT_k)}}\right) = \frac{1}{M^2T_k^2}
$
as the analysis of \eqref{eq:proof-single-second-bound-eta}.
Thus we have $\|E_k^1\|_{\infty}\lesssim \frac{\|h^{\star}\|_{\mathsf{sp}}}{M^2T_k^2}$, $\|E_k^2\|_{\infty}\lesssim \frac{k\|h^{\star}\|_{\mathsf{sp}}}{M^2T_k^2}$, and $\|E_{k,j}^3\|_{\infty}\lesssim \frac{k\|h^{\star}\|_{\mathsf{sp}}}{M^2T_k^2}$ for $j<k$.
Then we obtain
\begin{align}\label{eq:proof-fl-second-Delta_k_decomp_3}
\Delta_{k+1} 
&=\frac{k\eta_k}{k+1}\sum_{i=0}^{N_k-1}(1-\eta_k)^{N_k-1-i}P(V_{k} - V_k^{\star}) \notag\\
&\quad + \frac{k\eta_k}{M(k+1)}\sum_{m=1}^M\sum_{i=0}^{N_k-1}(1-\eta_k)^{N_k-1-i}(P_{k,i}^m-P)V_{k} + O\left(\frac{k\|h^{\star}\|_{\mathsf{sp}}}{M^2T_k^2}\right).
\end{align}

\paragraph{Step 2: bounding the decomposed errors.}
Repeating the argument in \eqref{eq:Delta_k_decomp_3}, we obtain
\begin{align}\label{eq:proof-fl-second-Delta}
& \Delta_{k+1} \nonumber \\
&\le \underbrace{\frac{1}{k+1} \prod_{l=1}^k(\alpha_lP^{\pi_l})\Delta_1}_{E_{k,k,0}^3} + \sum_{j=1}^k\prod_{l=j+1}^k(\alpha_lP^{\pi_l})\frac{j}{k+1}\underbrace{\frac{\eta_j}{M}\sum_{m=1}^M\sum_{i=0}^{N_j-1}(1-\eta_j)^{N_j-1-i}(P_{j,i}^m-P)V_{j}}_{E_{k,k,j}^3} + 
O\left(\frac{\|h^{\star}\|_{\mathsf{sp}}}{(k+1)c_N^2}\right).
\end{align}
Since $\|E_{k,k,0}^3\|_{\infty}$ can be bounded by $\widetilde{O}(\frac{\|h^{\star}\|_{\mathsf{sp}}}{k+1})$, as shown in \eqref{eq:proof-single-second-bound-Ekk03} for the single-agent case, it suffices to focus on establishing the bound for the second error term, $E_{k,k,j}^3$.
To this end, we compute the variance of $E_{k,k,j}^3$ as described in \eqref{eq:proof-single-second-varEkkj3} for the single-agent scenario:
\begin{align*}
\mathsf{Var}\left(E_{k,k,j}^3\right) 
&\le \frac{\eta_j}{M}\left(\|\Delta_j\|_{\infty} + \frac{\|h^{\star}\|_{\mathsf{sp}}}{j}\right)^2.
\end{align*}
By applying Bernstein's inequality, with probability at least $1-\delta/(2j^2)$, we have 
\begin{align}\label{eq:proof-fl-second-bound-Ekkj3}
E_{k,k,j}^3
& \lesssim \sqrt{\frac{\eta_j}{M}}\left(\|\Delta_j\|_{\infty} + \frac{\|h^{\star}\|_{\mathsf{sp}}}{j}\right)\sqrt{\log\frac{SAj}{\delta}} + \frac{\eta_j}{M}\|V_j-J^{\star}\|_{\infty}\log\frac{2SAj^2}{\delta}\notag\\
&\overset{\text{(i)}}{\lesssim} \sqrt{\frac{\log T_j}{MN_j}}\left(\|\Delta_j\|_{\infty} + \frac{\|h^{\star}\|_{\mathsf{sp}}}{j}\right)\sqrt{\log\frac{SAj}{\delta}}\notag\\
&\overset{\text{(ii)}}{\lesssim}  \frac{1}{j\log (j+1)}\left(\|\Delta_j\|_{\infty} + \frac{\|h^{\star}\|_{\mathsf{sp}}}{j}\right)\frac{1}{\sqrt{c_N}},
\end{align}
where (i) arises from the definition of $\eta_j$ (cf. \eqref{eq:fed-para-group-2}):
\begin{align*}
\eta_j = \frac{1}{1+\frac{N_j}{4\log (MT_j)}} = \frac{1}{1+\frac{N_j}{4\log(MT_j)}}\le \frac{4\log (MT_j)}{N_j},
\end{align*}
and (ii) arises from \eqref{eq:proof-fl-second-claim}.
By substituting \eqref{eq:proof-fl-second-bound-Ekkj3} into \eqref{eq:proof-fl-second-Delta_k_decomp_3}, we derive the recursive formula presented in \eqref{eq:proof-single-second-Delta-upper} and \eqref{eq:proof-singleagent-recursion}.

\paragraph{Step 3: solving recursion.}
Repeating argument in Step 3 in Section \ref{subsec:proof-thm-single-secondgroup}, 
we have
\begin{align}
\|Q_{K,N_K}^m - J^{\star}\|_{\infty} \le \|Q_{K,N_K}^m - Q_K^{\star}\|_{\infty} + \|J^{\star} - Q_K^{\star}\|_{\infty} \le \left\|\Delta_{K}\right\|_{\infty} + \frac{\|h^{\star}\|_{\mathsf{sp}}}{K}\lesssim \frac{\|h^{\star}\|_{\mathsf{sp}}\log K}{K},
\end{align}
which is consistent with the result in single-agent setting.
The sample complexity and the number of communications are obtained immediately.

\paragraph{Proof of \eqref{eq:proof-fl-second-claim-1} and \eqref{eq:proof-fl-second-claim}.}

Let
$$
i_{\min} = \arg \min \left\{i: M\log\left(M\log\frac{SA}{\delta}\right)\log(i+1) \ge \frac{i^2}{M}\log^5(i+1)\log^3\left(\frac{SA}{\delta}\right)\right\}.
$$
We then have $N_k = c_N\log\left(M\log\frac{SA}{\delta}\right)\log(k+1)$ for $j< i_{\min}$ and $N_k = c_N\frac{k^2}{M}\log^5(k+1)\log^3\frac{SA}{\delta}$ for $j\ge i_{\min}$.
Next, we proceed by proving \eqref{eq:proof-fl-second-claim-1} for the cases $k< i_{\min}$ and $k\ge i_{\min}$ separately.
For the case $k< i_{\min}$, assuming $c_N$ is sufficiently large, we have
\begin{align}\label{eq:proof-fl-second-claim-case1}
N_k &= c_N\log\left(M\log\frac{SA}{\delta}\right)\log(k+1)
\ge \log c_N + 2\log M + \log\log\log\frac{SA}{\delta} + 2\log(k+1)\notag\\
&\ge \log\left(c_NkM\log\left(M\log\frac{SA}{\delta}\right)\log(k+1)\right) = \log(MT_k).
\end{align}
In the case where $k\ge i_{\min}$, we have the following inequality:
\begin{align}
N_k &=\frac{c_N}{M}k^2\log^5(k+1)\log^3\frac{SA}{\delta} \ge c_N\log\left(M\log\frac{SA}{\delta}\right)\log(k+1)\notag\\
&\ge \log c_N + 8\log(k+1) + 3\log\log\frac{SA}{\delta}\notag\\
&\ge \log\left(c_Nk^3\log^5(k+1)\log^3\frac{SA}{\delta}\right)
=\log(MkN_k) \ge \log(MT_k).
\end{align}

Similarly, \eqref{eq:proof-fl-second-claim} is derived from the following expression:
\begin{align}
MN_k \ge c_Nk^2\log^5(k+1)\log^3\frac{SA}{\delta} \ge c_Nk^2\log^2(k+1)\log\frac{SA}{\delta}\left(\log^3(k+1)\log^2\frac{SA}{\delta}\right),
\end{align}
and
\begin{align*}
\log(MT_k)
&\le\log(kMN_k) 
= \log\left(c_NkM\log\left(M\log\frac{SA}{\delta}\right)\log(k+1)\right)\notag\\
&\lesssim \log(c_NM) + \log k + \log\log\log\frac{SA}{\delta} \le 
M\log^3(k+1)\log^2\frac{SA}{\delta}.
\end{align*}

%% file: analysis_policy.tex
\section{Analysis for optimal policy learning (Theorem \ref{thm:avg-policy-fed})}
\label{sec:analysis_policy}

We first decompose the estimation error of the optimal policy as
\begin{align}\label{eq:proof-policy-decomp-J}
0\le J^{\star} - J^{\widehat{\pi}} \le J^{\star} - V_{\gamma_K}^{\star} + V_{\gamma_K}^{\star} - V_{\gamma_K}^{\widehat{\pi}} + V_{\gamma_K}^{\widehat{\pi}} - J^{\widehat{\pi}}.
\end{align}
According to Lemma 6 - Lemma 8 in \citet{wang2022near}, we have
\begin{align}
\|J^{\star} - V_{\gamma_K}^{\star}\|_{\infty}&\le 4(1-\gamma_K)\|h^{\star}\|_{\mathsf{sp}},\label{eq:proof-policy-temp-1}\\ \|J^{\widehat{\pi}} - V_{\gamma_K}^{\widehat{\pi}}\|_{\infty}&\le \|V_{\gamma_K}^{\widehat{\pi}}\|_{\mathsf{sp}} \le \|V_{\gamma_K}^{\widehat{\pi}} - V_{\gamma_K}^{\star}\|_{\infty} + \|V_{\gamma_K}^{\star}\|_{\mathsf{sp}}
\le 
4(1-\gamma_K)\|h^{\star}\|_{\mathsf{sp}} + \|V_{\gamma_K}^{\star} - V_{\gamma_K}^{\widehat{\pi}}\|_{\infty}.\label{eq:proof-policy-temp-2}
\end{align}
Moreover, by definition, we control the estimation error $\|V_{\gamma_K}^{\star} - V_{\gamma_K}^{\widehat{\pi}}\|_{\infty}$ by using the estimation error for $Q^{\star}_{\gamma_K}$ as follows:
\begin{align}
\|V_{\gamma_K}^{\star} - V_{\gamma_K}^{\widehat{\pi}}\|_{\infty} 
&= \max_s\frac{Q_{\gamma_K}^{\star}(s,\pi^{\star}_{\gamma_K}(s)) - Q_{\gamma_K}^{\star}(s,\widehat{\pi}(s))}{1-\gamma_K} \notag\\
&= \frac{1}{1-\gamma_K}\max_s\big(Q_{\gamma_K}^{\star}(s,\pi^{\star}_{\gamma_K}(s)) - Q_{K,N_K}(s,\pi^{\star}_{\gamma_K}(s)) \notag\\
&\quad \quad + Q_{K,N_K}(s,\pi^{\star}_{\gamma_K}(s)) - Q_{K,N_K}(s,\widehat{\pi}(s)) + Q_{K,N_K}(s,\widehat{\pi}(s)) - Q_{\gamma_K}^{\star}(s,\widehat{\pi}(s))\big)\notag\\
&\overset{\text{(i)}}{\le} \frac{2\|Q_{\gamma_K}^{\star} - Q_{K,N_K}\|_{\infty}}{1-\gamma_K},\label{eq:proof-policy-temp-3}
\end{align}
where (i) holds since $Q_{K,N_K}(s,\pi^{\star}_{\gamma_K}(s)) - Q_{K,N_K}(s,\widehat{\pi}(s))\le 0$ by the definition of $\widehat{\pi}$.
Substituting \eqref{eq:proof-policy-temp-1}, \eqref{eq:proof-policy-temp-2}, and \eqref{eq:proof-policy-temp-3} into \eqref{eq:proof-policy-decomp-J}, we have
\begin{align*}
\|J^{\star} - J^{\widehat{\pi}}\|_{\infty}
&\le \|J^{\star} - V_{\gamma_K}^{\star}\|_{\infty} + \|V_{\gamma_K}^{\star} - V_{\gamma_K}^{\widehat{\pi}}\|_{\infty} + \|V_{\gamma_K}^{\widehat{\pi}} - J^{\widehat{\pi}}\|_{\infty}\notag\\
&\le 8(1-\gamma_K)\|h^{\star}\|_{\mathsf{sp}} + 2\|V_{\gamma_K}^{\star} - V_{\gamma_K}^{\widehat{\pi}}\|_{\infty}
\le 8(1-\gamma_K)\|h^{\star}\|_{\mathsf{sp}} + \frac{4\|Q_{\gamma_K}^{\star} - Q_{K,N_K}\|_{\infty}}{1-\gamma_K}.
\end{align*}
Following similar analysis as that in federated setting, we claim that
\begin{align}\label{eq:proof-policy-claim-Qlearning}
\|Q_{\gamma_K}^{\star} - Q_{K,N_K}\|_{\infty}\lesssim \sqrt{\frac{\log\frac{SAMT_K}{\delta}}{(1-\gamma_K)MN_K}}\|h^{\star}\|_{\mathsf{sp}}\log(N_K M),
\end{align}
whose proof is postponed to the end of this section.
Recalling the choices of $\gamma_K$, we have
\begin{align*}
J^{\star} - J^{\widehat{\pi}}\lesssim \frac{\|h^{\star}\|_{\mathsf{sp}}}{(MN_K)^{1/5}} + \sqrt{\frac{\log\frac{SAMT_K}{\delta}}{(1-\gamma_K)^3MN_K}}\|h^{\star}\|_{\mathsf{sp}}\log(N_K M)\lesssim \frac{\|h^{\star}\|_{\mathsf{sp}}}{(MN_K)^{1/5}}\log^{\frac12}\frac{SAMT_K}{\delta}\log(N_K M)
\end{align*}
Provided that
$$
N_K\gtrsim \frac{\|h^{\star}\|_{\mathsf{sp}}^5}{\varepsilon^5 M}\log^5(N_KM)\log^{\frac52}\frac{SAMT_K}{\delta},
$$
we have $J^{\star} - J^{\widehat{\pi}}\le \varepsilon$.
Recalling that $N_k = c_N 2^k$, we have $T_K = \sum_{k=1}^K N_k \le 2 N_K$ and complete the proof.

For the number of communication rounds, we have
\begin{align*}
|\mathcal{C}_k| = \frac{4\log(1-\gamma_k)}{\log((1+\gamma_k)/2)} + 1 \le \frac{8\log(\frac{1}{1-\gamma_k})}{1-\gamma_k} = \frac{8}{5}(N_kM)^{1/5}\log(N_kM)\le \frac{8}{5}(N_kM)^{1/5}\log(T_KM).
\end{align*}
Thus we have
\begin{align*}
\sum_{k=1}^K |\mathcal{C}_k| \le \frac{8\log(T_KM)}{5}\sum_{k=1}^K(N_kM)^{1/5} \lesssim (N_KM)^{1/5}\log(T_KM).
\end{align*}

\paragraph{Proof of \eqref{eq:proof-policy-claim-Qlearning}.}
According to the update rule of $Q_{k,t}$, we have
\begin{align*}
Q_{k,t+1}^m - Q_{\gamma_k}^{\star} 
& = \prod_{j=1}^t(1-\eta_{k,j})(Q_{k,0} - Q_{\gamma_k}^{\star}) + \gamma_k\sum_{i=1}^t\eta_{k,i}\prod_{j=i+1}^t(1-\eta_{k,j})(P_{k,i}^mV_{k,\iota(k,i)} - PV_{\gamma_{k}}^{\star})\notag\\
& = \prod_{j=1}^t(1-\eta_{k,j})(Q_{k,0} - Q_{\gamma_k}^{\star}) + \gamma_k\sum_{i=1}^{\iota(k,t)}\eta_{k,i}\prod_{j=i+1}^t(1-\eta_{k,j})(P_{k,i}^mV_{k,\iota(k,i)} - PV_{\gamma_{k}}^{\star}) \notag\\
&\quad + \gamma_k\sum_{i=\iota(k,t)+1}^{t}\eta_{k,i}\prod_{j=i+1}^t(1-\eta_{k,j})(P_{k,i}^mV_{k,\iota(k,i)} - PV_{\gamma_{k}}^{\star}).
\end{align*}
Denote by $t_h$ the $h$-th smallest value in the set $\mathcal{C}_k$ for $h\ge 1$, and let $t_0 = 0$.
Note that for $h\ge 2$, we have
\begin{align}
\prod_{j=t_{h-1}+1}^{t_h}(1-\eta_{k,j}) \le \exp\left(-\sum_{j=t_{h-1}+1}^{t_h}\eta_{k,j}\right) \le \exp\left(-(t_h - t_{h-1})\eta_{k,t_h}\right).
\end{align}
Considering that for $h\ge 2$, we have $t_h - t_{h-1} = (1-\gamma_k)t_h/2$, and 
\begin{align*}
\eta_{k,t_h} = \left(1+\frac{t_h(1-\gamma_k)}{8\log(MN_k)}\right)^{-1},\quad \forall h \ge 1.
\end{align*}
If $t_h\ge 8\log(MN_k)/(1-\gamma_k)$, then we have $\eta_{k,t_h} \ge \frac{4\log(MN_k)}{t_h(1-\gamma_k)}$ and
\begin{align}\label{eq:proof-policy-prodt}
\prod_{j=t_{h-1}+1}^{t_h}(1-\eta_{k,j}) \le \exp\left(-\frac{(1-\gamma_k)t_h}{2}\frac{4\log(MN_k)}{t_h(1-\gamma_k)}\right) \le \frac{1}{M^2N_k^2}.
\end{align}
We are now ready to introduce an error sequence
$$
\Delta_{k,h}\coloneqq \|Q_{k,t_h} - Q_{\gamma_k}^{\star}\|_{\infty},
$$
where $Q_{k,t_h}$ denotes the result after the $h$-th communication round, shared identically among all agents.
From \eqref{eq:proof-policy-prodt}, and noting that $Q_{k,t}, Q_{\gamma_k}^{\star} \le 1$, we obtain
\begin{align}\label{eq:proof-policy-delta-recur1}
\Delta_{k,h} \le \frac{2}{M^2N_k} + \|\mathcal{E}_{k,h}^{(1)}\|_{\infty} + \|\mathcal{E}_{k,h}^{(2)}\|_{\infty}, \quad h \ge 1,
\end{align}
where
\begin{align}
\mathcal{E}_{k,h}^{(1)}
&\coloneqq \frac{\gamma_k}{M}\sum_{m=1}^M\sum_{i=t_{h-1}+1}^{t_h}\eta_{k,i}\prod_{j=i+1}^{t_h}(1-\eta_{k,j})(P_{k,i}^m - P)V_{\gamma_{k}}^{\star}),\\
\mathcal{E}_{k,h}^{(2)}
&\coloneqq \frac{\gamma_k}{M}\sum_{m=1}^M\sum_{i=t_{h-1}+1}^{t_h}\eta_{k,i}\prod_{j=i+1}^{t_h}(1-\eta_{k,j})P_{k,i}^m(V_{k,t_{h-1}} - V_{\gamma_{k}}^{\star}.
\end{align}
For $h\ge 2$, direct calculation yields
\begin{align*}
\frac{\gamma_k}{M}\sum_{m=1}^M\sum_{i=t_{h-1}+1}^{t_h}\eta_{k,i}\prod_{j=i+1}^{t_h}(1-\eta_{k,j}) & \le \gamma_k,\notag\\
\frac{\gamma_k^2}{M^2}\sum_{m=1}^M\sum_{i=t_{h-1}+1}^{t_h}\eta_{k,i}^2 
&\le \frac{\gamma_k^2}{M}\sum_{i=t_{h-1}+1}^{t_h}\eta_{k,i}^2\le \frac{\gamma_k^2(t_h - t_{h-1})}{M} \eta_{k,t_{h-1}}^2\notag\\
& \le \frac{\gamma_k^2(1-\gamma_k)t_h}{2M}\frac{64\log^2(MN_k)}{t_{h-1}^2(1-\gamma_k)^2} \lesssim \frac{\gamma_k^2\log^2(MN_k)}{(1-\gamma_k)Mt_h},
\end{align*}
where we have used $t_h / t_{h-1} \le 2$.
Recall that $\|V_{\gamma_k}^{\star}\|_{\mathsf{sp}} \lesssim (1-\gamma_k)\|h^{\star}\|_{\mathsf{sp}}$.
By Hoeffding's inequality, with probability at least $1-\delta$, we have
\begin{align*}
\left\|\mathcal{E}_{k,h}^{(1)}\right\|_{\infty}\lesssim \gamma_k\sqrt{\frac{\log\frac{SAMT_K}{\delta}}{(1-\gamma_k)Mt_h}}(1-\gamma_k)\|h^{\star}\|_{\mathsf{sp}} \log(MN_k)\asymp \gamma_k\sqrt{\frac{(1-\gamma_k)\log\frac{SAMT_K}{\delta}}{Mt_h}}\|h^{\star}\|_{\mathsf{sp}} \log(MN_k)
\end{align*}
Moreover, we have $\left\|\mathcal{E}_{k,h}^{(2)}\right\|_{\infty} \le \gamma_k \Delta_{k,h-1}$.
Substituting these into \eqref{eq:proof-policy-delta-recur1}, for $h\ge 2$, we obtain
\begin{align}
\Delta_{k,h} 
&\le \gamma_k\Delta_{k,h-1} + C\gamma_k\|h^{\star}\|_{\mathsf{sp}}\sqrt{\frac{(1-\gamma_k)\log\frac{SAMT_K}{\delta}}{Mt_h}} \log(MN_k)\notag\\
&\le \gamma_k^{h-1}\Delta_{k,1} + C\gamma_k\|h^{\star}\|_{\mathsf{sp}}\log(MN_k)\sqrt{\frac{(1-\gamma_k)\log\frac{SAMT_K}{\delta}}{M}}\sum_{i=1}^h\gamma_k^{h-i}\sqrt{\frac{1}{t_i}}\notag\\
&\le \gamma_k^{h-1} + C\gamma_k\|h^{\star}\|_{\mathsf{sp}}\log(MN_k)\sqrt{\frac{(1-\gamma_k)\log\frac{SAMT_K}{\delta}}{M}}\sum_{i=1}^h\left(\frac{2\gamma_k}{1+\gamma_k}\right)^{h-i}\sqrt{\frac{1}{t_h}}\notag\\
&\le \gamma_k^{h-1} + C\gamma_k\|h^{\star}\|_{\mathsf{sp}}\log(MN_k)\sqrt{\frac{(1-\gamma_k)\log\frac{SAMT_K}{\delta}}{M}}\sum_{i=1}^h\left(\frac{1+\gamma_k}{2}\right)^{h-i}\sqrt{\frac{1}{t_h}}.
\end{align}
Let
$$
H = \left\lceil\frac{\log((1-\gamma_k)^2)}{\log(\frac{1+\gamma_k}{2})}\right\rceil.
$$
Then we have
$$
\Delta_{k,H} \le \gamma_k^{H-1} + 2C\gamma_k\|h^{\star}\|_{\mathsf{sp}}\log(MN_k)\sqrt{\frac{\log\frac{SAMT_K}{\delta}}{(1-\gamma_k)MN_k}}\lesssim \|h^{\star}\|_{\mathsf{sp}}\log(MN_k)\sqrt{\frac{\log\frac{SAMT_K}{\delta}}{(1-\gamma_k)MN_k}},
$$
which completes the proof.